\pdfoutput=1

\documentclass{article}

\usepackage[eandd, final]{neurips_2026}

\usepackage[utf8]{inputenc}
\usepackage[T1]{fontenc}
\usepackage{hyperref}
\hypersetup{hidelinks}
\usepackage{url}
\usepackage{booktabs}
\usepackage{graphicx}
\usepackage{array}
\usepackage{tabularx}
\usepackage{amsfonts}
\usepackage{amsmath}
\usepackage{amssymb}        %
\usepackage{nicefrac}
\usepackage[expansion=false]{microtype}  %
\usepackage{xcolor}
\usepackage{float}                          %
\usepackage{placeins}                       %
\usepackage{fancyvrb}                       %
\usepackage{fvextra}                         %
\usepackage{textcomp}                        %
\usepackage{etoc}                            %
\usepackage[capitalize,noabbrev]{cleveref}   %

\newcommand{\AgriManager}{\textsc{AgriManager}}
\newcolumntype{Y}{>{\raggedright\arraybackslash}X}

\title{\raisebox{-0.22em}{\includegraphics[height=1.25em]{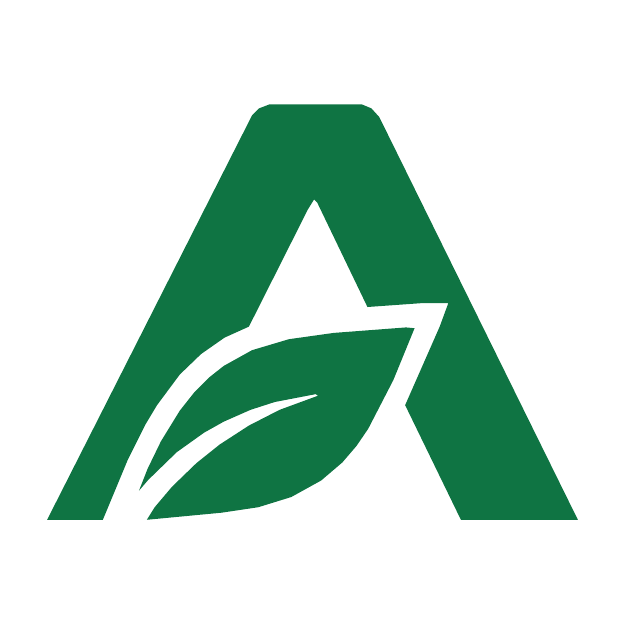}}\,\AgriManager: A Framework and Benchmark for LLM-RL Generalization in Agricultural Management}

\author{%
  Chi Gui$^{1}$ \quad
  Junwen Zheng$^{1}$ \quad
  Ritwik Nigam$^{1}$ \quad
  Qianlan Yang$^{1}$ \quad
  Meagan Lang$^{1}$ \\
  \bfseries
  Vardhan Dongre$^{1}$ \quad
  Jeremiah Barr$^{2}$ \quad
  Yu-Xiong Wang$^{1}$ \quad
  Vikram S. Adve$^{1}$ \\[4pt]
  \mdseries
  $^{1}$University of Illinois Urbana-Champaign \quad
  $^{2}$Corteva Agriscience \\
  \texttt{\{chigui2,vadve\}@illinois.edu}
}

\begin{document}

\maketitle
\etocdepthtag.toc{mtmain}

\begin{abstract}

Agricultural management is heterogeneous along two axes. The first is distributional shift within a fixed task interface: weather, crops, and prices vary across years and locations while sensors, actions, and objective stay fixed. The second is structural shift across interfaces: different farms expose different observed variables, available actions, and management objectives. We formalize the second axis as \emph{schema shift}: variation in the policy-facing schema $\Sigma = (\Sigma_O, \Sigma_A, \Sigma_R)$ between training and deployment. A fixed-interface neural baseline cannot run once a shift changes its input layout or action encoding and has no objective conditioning for a $\Sigma_R$ shift, while prompt-conditioned LLM policies re-read $\Sigma$ from each episode's prompt. We test whether LLM-RL generalizes across both axes with \emph{AgriManager}, to our knowledge the first framework that trains LLM policies with RL across multiple process-based crop simulators, unifying DSSAT, WOFOST, and Cycles under prompt-based schema specification with joint multi-objective, multi-environment training. We design a three-tier generalization ladder: Tier 1 within-schema distributional shift (weather, crop, price), Tier 2 single-axis schema shift along $\Sigma_O / \Sigma_A / \Sigma_R$, Tier 3 compositional schema shift across simulators, all scored on a frozen held-out test suite with three training seeds per method. Within fixed schemas, Qwen3-4B trained by RL with explicit reasoning exceeds neural-network PPO trained with the same interaction budget; without reasoning, training stalls near no-action where the reward penalizes input use. Under single-axis schema shift, $\Sigma_O$ transfer needs decision-critical state, $\Sigma_A$ results are consistent with joint action-menu coverage, and with reasoning a policy trained on yield alone transfers to unseen objectives. Under compositional schema shift, single-source training mostly hurts zero-shot transfer across simulators, and a unified policy trained only on other combinations of the same components recovers about 86\% of the gain of a specialist trained on the target combination on DSSAT and stays below the specialists on the WOFOST and Cycles profit targets; closing this gap remains open. These results give an initial baseline for unified prompt-conditioned control in simulated agricultural management.

\noindent Project page: \url{https://agrimanager-framework.github.io/}

\end{abstract}

\section{Introduction}
\label{sec:introduction}

Crop management decisions are made each growing season and affect both farm productivity and environmental outcomes. Work on these decisions has long drawn on process-based crop simulation and rule- or optimization-based scheduling~\citep{jones2017brief}. Recent machine-learning work in agriculture has concentrated more on crop sensing and yield prediction than on the management decisions themselves~\citep{kamilaris2018dlag,vanklompenburg2020yield}. Reinforcement learning targets the decision directly, casting management as sequential control under uncertainty. New agricultural RL environments make this formulation executable across crop simulators~\citep{overweg2021cropgym,gautron2022gymdssat,turchetta2022cyclesgym,solow2025wofostgym}. The open question is whether such policies generalize across the heterogeneity inherent to crop management.

The heterogeneity an agricultural RL policy faces decomposes into two qualitatively different axes. The first is \emph{distributional shift within a fixed interface}: weather, soil, crop variety, or prices change while the observed variables, available actions, and management objective stay the same. This is the standard setting in the RL generalization literature, studied through contextual MDPs~\citep{hallak2015contextual}, procedural-generation benchmarks~\citep{cobbe2020procgen}, and domain randomization~\citep{tobin2017dr}, and surveyed~\citep{kirk2023survey,zhao2020sim2real}. The second is \emph{structural shift across interfaces}: different farms expose different observed variables, support different management actions, or optimize different objectives, so the policy-facing interface itself changes. Prior work has studied parts of this second axis: transfer across observation feature spaces~\citep{sun2022obsfeature} or heterogeneous action spaces~\citep{sivakumar2024heterogeneous}, differences in MDP components surveyed for transfer RL~\citep{zhu2023transfersurvey}, and, for LLM agents, changes in observation spaces, action interfaces and tool schemas~\citep{xi2026rlgeneralization,wu2026openagent,liu2026actionspace}. Joint changes across observations, actions and objectives, evaluated together in one benchmark, remain less studied. We name this broader structural axis \emph{schema shift}, orthogonal to the distributional axis above. In agricultural RL specifically, to our knowledge neither axis has been systematically evaluated.

Current agricultural RL benchmarks, however, treat the policy-facing interface as a fixed property of the simulator: each environment binds one observation set, one action menu, and one reward formulation to one underlying simulator~\citep{gautron2022gymdssat,solow2025wofostgym,turchetta2022cyclesgym,overweg2021cropgym}. The schema is therefore not a benchmark variable, and structural generalization across observed variables, action menus, or reward formulations falls outside what current setups can measure. Prompt-conditioned LLM policies~\citep{szot2024llarp,ahn2022saycan,yao2023react} now offer a different policy class: they ingest the schema directly from text at inference time, so a single trained policy can in principle operate across schemas that would otherwise each require their own neural architecture or adapter. Yet no agricultural benchmark currently supports such policies: the closest related work uses a language model only as a text encoder for a downstream DQN~\citep{wu2024newagronomists}, and broader LLM--RL integrations remain at the survey level~\citep{chen2024llmrlcropsurvey}. The infrastructure to test this policy class is therefore missing.

We introduce \emph{AgriManager}, a framework and benchmark that make the policy-facing schema an explicit, variable axis of agricultural RL (Sections~\ref{sec:framework} and~\ref{sec:benchmark}). Table~\ref{tab:related-work-comparison} contrasts it with representative prior work, none of which jointly covers the agricultural domain, multi-simulator integration, schema shift, and joint multi-objective, multi-environment LLM-RL training; Appendix~\ref{app:related-work} gives the per-thread discussion.

\textbf{Contributions.} We (1)~introduce the AgriManager framework, which unifies three crop simulators (DSSAT, WOFOST, and Cycles) under one prompt-based schema interface and trains one prompt-conditioned policy jointly across simulators and objectives, with fixed-interface neural baselines (Section~\ref{sec:framework}); (2)~construct a three-tier benchmark of eight settings spanning within-schema distributional shift, schema shift targeted at one axis, and compositional schema shift across simulators, scored on a frozen held-out test suite against no-action and a scripted agronomic reference and released with its manifests, seeds, and evaluation scripts (Section~\ref{sec:benchmark}); and (3)~find that, with explicit reasoning, LLM-RL on a Qwen backbone exceeds neural-network PPO trained with the same interaction budget under distributional shift, shows an axis-specific profile under schema shift, and yields a unified policy that, trained only on other component combinations, recovers about 86\% of a target-trained specialist's gain on DSSAT and stays below the specialists on the WOFOST and Cycles profit targets, while single-source training mostly hurts zero-shot cross-simulator transfer (Sections~\ref{sec:tier1}--\ref{sec:tier3}).

\newcommand{\rwYes}{\ensuremath{\checkmark}}
\newcommand{\rwNo}{\ensuremath{-}}

\begin{table}[t]
\centering
\small
\setlength{\tabcolsep}{4pt}
\caption{Comparison of \AgriManager{} against representative prior work along five axes: (1)~training NN policies with RL on the environment (PPO, DQN, etc.); (2)~training a prompt-conditioned LLM policy with RL; (3)~integration of multiple simulator backends; (4)~explicit study of schema-shift generalization; (5)~joint multi-objective and multi-environment training. \emph{eval.}: prompted LLM agents are evaluated, without RL training. The row marked $(m)$ is a method paper, included to position our LLM-RL contribution against the most directly related prior agricultural-RL method.}
\label{tab:related-work-comparison}
\begin{tabular}{lccccc}
\toprule
                                                   & NN-RL    & prompt-     & multi-      & schema           & multi-obj +   \\
                                                   & training & LLM-RL      & simulator   & shift            & multi-env     \\
\midrule
gym-DSSAT~\citep{gautron2022gymdssat}               & \rwYes   & \rwNo       & \rwNo       & \rwNo            & \rwNo \\
WOFOSTGym~\citep{solow2025wofostgym}                & \rwYes   & \rwNo       & \rwNo       & \rwNo            & \rwNo \\
CyclesGym~\citep{turchetta2022cyclesgym}            & \rwYes   & \rwNo       & \rwNo       & \rwNo            & \rwNo \\
CropGym~\citep{overweg2021cropgym}                  & \rwYes   & \rwNo       & \rwNo       & \rwNo            & \rwNo \\
\citet{wu2024newagronomists}~$(m)$                  & \rwYes   & \rwNo       & \rwNo       & \rwNo            & \rwNo \\
ProcGen~\citep{cobbe2020procgen}                    & \rwYes   & \rwNo       & \rwNo       & \rwNo            & \rwNo \\
\citet{sun2022obsfeature}                           & \rwYes   & \rwNo       & \rwNo       & $\Sigma_O$ only  & \rwNo \\
AgentBoard~\citep{ma2024agentboard}                 & \rwNo    & eval.       & \rwNo       & \rwNo            & \rwNo \\
WebArena~\citep{zhou2024webarena}                   & \rwNo    & eval.       & \rwNo       & \rwNo            & \rwNo \\
\textbf{\AgriManager{} (ours)}                      & \rwYes   & \rwYes      & \rwYes      & \rwYes           & \rwYes \\
\bottomrule
\end{tabular}
\end{table}

\section{The AgriManager framework}
\label{sec:framework}

\begin{figure}[!htbp]
  \centering
  \includegraphics[width=\textwidth]{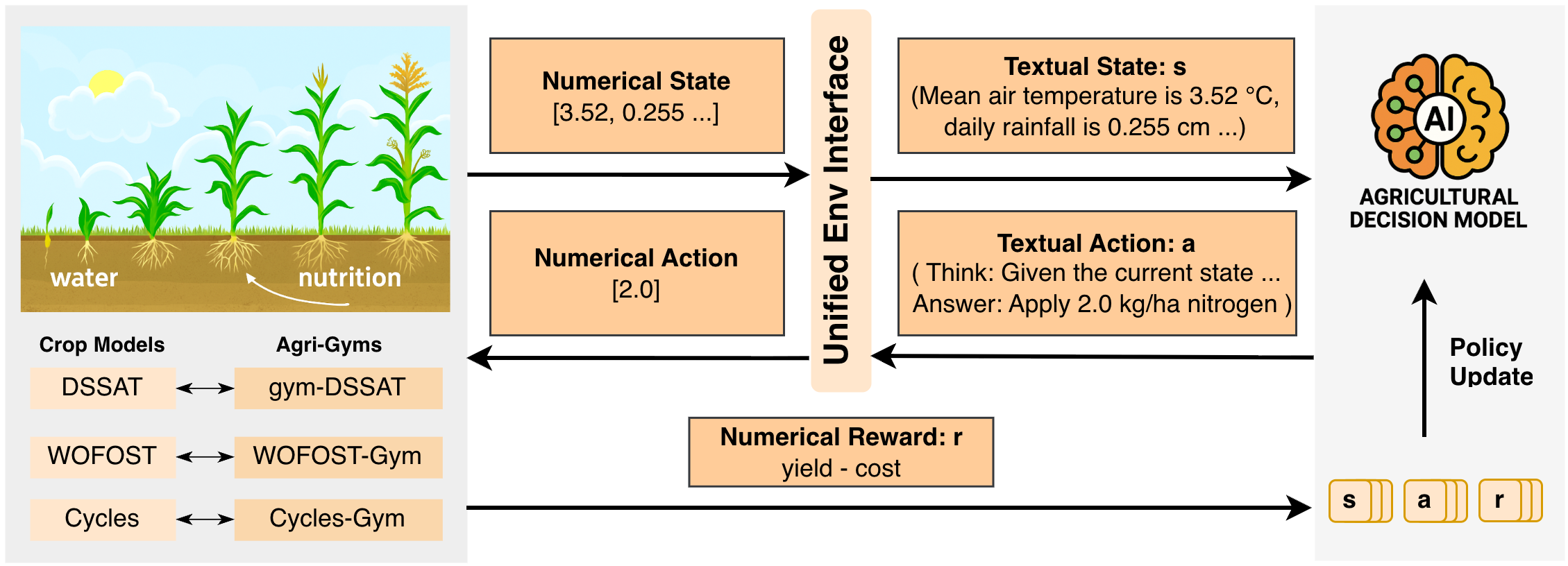}
  \caption{The AgriManager framework architecture. The ``Agri-Gyms'' layer wraps the three crop simulators; the environment adapters and Unified Env Interface translate the numerical state-action representation into the text consumed by the LLM decision model; rollouts $(s, a, r)$ drive policy updates.}
  \label{fig:agrimanager-architecture}
\end{figure}

We introduce \emph{AgriManager}, to our knowledge the first framework that trains LLM policies with RL across multiple process-based crop simulators. The framework unifies the DSSAT, WOFOST, and Cycles simulators under a schema-aware text interface, declares the policy-facing schema $\Sigma$ to the policy through a structured prompt template (Appendix~\ref{app:prompts}), and supports multi-turn rollout with joint multi-objective, multi-environment training (Section~\ref{sec:framework-training}). The crop simulators are accessed through their gym wrappers (gym-DSSAT~\citep{gautron2022gymdssat}, WOFOSTGym~\citep{solow2025wofostgym}, CyclesGym~\citep{turchetta2022cyclesgym}), and we name each environment after its simulator: DSSAT, WOFOST, and Cycles. Figure~\ref{fig:agrimanager-architecture} shows the architecture; adapters for additional simulators follow a documented contract (Appendix~\ref{app:adapter-contract}).

\subsection{Decision process and schema shift}
\label{sec:framework-pomdp}

We frame an agricultural management task as a finite-horizon Partially Observable Markov Decision Process (POMDP)~\citep{kaelbling1998planning} $\mathcal{M} = (\mathcal{S}, \mathcal{O}, \mathcal{A}, P, E, R, \rho_0, \gamma, H)$, with transition kernel $P$, observation kernel $E$, reward $R$, and horizon $H$ (a crop season or a multi-year rotation).

The policy-facing interface of $\mathcal{M}$ is described by a \emph{schema} $\Sigma = (\Sigma_O, \Sigma_A, \Sigma_R)$ that specifies what the policy observes, what actions it may take, and what objective it is optimizing. The observation schema $\Sigma_O$ is the finite list of labeled observed variables; both the number of variables and the labels themselves are part of the schema. The action schema $\Sigma_A$ is the menu of available management actions and their argument spaces. The reward schema $\Sigma_R$ specifies the form of the objective function. The remaining POMDP components $(\mathcal{S}, P, E, \rho_0)$ and the scalar parameters that calibrate the objective, such as prices, enter as episode context $z \sim q(z)$; $\Sigma$ is held fixed within an episode and may vary across episodes.

This separation induces two kinds of shift between training and test. A \emph{distributional shift} keeps the schema fixed while $q(z)$ changes ($\Sigma_{\text{train}} = \Sigma_{\text{test}}$, $q_{\text{train}}(z) \neq q_{\text{test}}(z)$); a \emph{schema shift} changes the policy-facing interface ($\Sigma_{\text{train}} \neq \Sigma_{\text{test}}$). Observation-schema, action-menu, and reward-formulation shifts correspond to changes in $\Sigma_O$, $\Sigma_A$, and $\Sigma_R$ respectively. Either kind can occur without the other; a change of $P$, as in sim-to-real transfer, counts here as distributional shift.

A fixed-interface neural policy, such as our PPO baseline, cannot run when a schema change alters its numerical input layout or action encoding, since its input layer and action head are sized to the training schema, and it has no objective conditioning, so a $\Sigma_R$ change leaves it optimizing the training reward. A neural policy over a shared, masked action vocabulary covers menus drawn from it (Appendix~\ref{app:bl-nnppo}). A prompt-conditioned LLM policy re-reads $\Sigma$ from the prompt of each new episode, so it can run under a change on any of the three schema axes.

\subsection{Tasks}
\label{sec:framework-tasks}

The framework includes two task families, both multi-turn instances of the formalism of Section~\ref{sec:framework-pomdp}.

\textbf{Crop-growth tasks}: Every few simulated days, the agent decides whether to fertilize or irrigate; the simulator advances soil and crop state until the next decision. The default objective is profit over a single growing season.

\textbf{Crop-planning tasks}: At the start of each year, the agent picks the year's crop and planting week; the simulator runs the full annual cycle, including off-season soil dynamics. The objective is to maximize gross revenue over a multi-year rotation.

Each simulator has a distinct role (Appendix~\ref{app:simulator-pool}). WOFOST, with over twenty crop trait files, six native action menus, and four objectives, hosts all three single-axis schema settings. DSSAT supplies independent crop--soil dynamics and is a source and target of cross-simulator transfer. Cycles supplies the only long-horizon task, the 19-year corn--soybean rotation of price-regime.

\subsection{Multi-turn training pipeline}
\label{sec:framework-training}

An episode is a multi-turn rollout. At step $t$, the policy $\pi_\theta$ receives the user prompt and emits a structured response containing the action $a_t \in \mathcal{A}$; we evaluate two variants, think and no-think, detailed in Appendix~\ref{app:prompts-response}. The action is sent to the environment, which returns the next observation and a scalar reward; the user prompt for step $t+1$ is reassembled around the new observation. The trajectory $\tau = \{(o_t, a_t, r_t)\}_{t=0}^{H-1}$ has return $R(\tau) = \sum_t \gamma^t r_t$.

Because the schema is declared at every step, a single LLM policy can be trained jointly across multiple objectives and multiple environments. Each training instance is a triple $(\Sigma, P, R)$ specifying the prompt-declared schema, the simulator dynamics, and the matching reward function; the \texttt{<management objective>} block of $\Sigma$ is paired with the environment-side reward $R$. Drawing $(\Sigma, P, R) \sim \mathcal{D}_{\text{train}}$ from a training distribution, we train $\pi_\theta$ to maximize
\begin{equation}
  \mathcal{J}(\theta) = \mathbb{E}_{(\Sigma, P, R) \sim \mathcal{D}_{\text{train}}} \, \mathbb{E}_{\tau \sim \pi_\theta(\cdot \mid \Sigma)} \bigl[R(\tau)\bigr].
  \label{eq:joint-objective}
\end{equation}
The Tier 3 unified policy (Section~\ref{sec:tier3}) instantiates this joint training.

Policy updates run on VeRL~\citep{sheng2024hybridflow}, with hyperparameters in Appendix~\ref{app:training-hparams}.

\section{The AgriManager benchmark}
\label{sec:benchmark}

The AgriManager benchmark decomposes the question of Section~\ref{sec:introduction}, whether prompt-conditioned LLM-RL generalizes across distributional and structural shift, into a three-tier generalization ladder of eight settings with a shared evaluation protocol.

\subsection{Three-tier generalization ladder and settings}
\label{sec:benchmark-ladder}

Table~\ref{tab:test-sets-overview} lists the eight settings. Tier 1 holds $\Sigma$ fixed and varies $q(z)$. Tier 2 targets one of $\Sigma_O$, $\Sigma_A$, or $\Sigma_R$ by editing one block of the prompt template (Appendix~\ref{app:prompts}); on the action axis, each menu also brings its own production wrapper (Appendix~\ref{app:benchmark-t2}). On identical scenarios, withholding the four resource-state fields takes the observation schema from 15 fields to 11, and a nitrogen-only menu takes the action menu from 17 actions to 5; the fixed-interface baseline cannot run on either. Tier 3 changes multiple components of $\Sigma$ together (compositional schema shift), and its cross-simulator cases also shift $P$, a co-occurring factor (Appendix~\ref{app:t3-inventory}).

\begin{table*}[t]
  \centering
  \small
  \setlength{\tabcolsep}{4.5pt}
  \renewcommand{\arraystretch}{1.24}
  \caption{The eight settings at a glance. \emph{Varies} names the schema or context component that changes between train and test (Section~\ref{sec:framework-pomdp}). \emph{Scenarios}: training configurations $\times$ training rows per configuration / held-out test rows (Appendix~\ref{app:benchmark-details}).}
  \label{tab:test-sets-overview}
  \begin{tabularx}{\textwidth}{@{}
                  >{\hsize=1.1\hsize}Y
                  >{\hsize=1.0\hsize}Y
                  >{\hsize=1.08\hsize}Y
                  >{\hsize=1.22\hsize}Y
                  >{\hsize=.60\hsize}Y
                  r@{}}
    \toprule
    Setting & Varies & Train & Test & Objective & Scenarios \\
    \midrule
    \multicolumn{6}{@{}l}{\textit{Tier 1: within-schema distributional shift}} \\
    \addlinespace[0.2em]
    \textbf{weather-regime} & weather $z$ & normal weather & drought, wet, hot, cold & profit & $2{\times}1{,}600$ / $1{,}280$ \\
    \textbf{crop-variety} & crop variety $z$ & 4 / 8 / 16 seen crops & 4 held-out crops & profit & $3{\times}1{,}600$ / $1{,}024$ \\
    \textbf{price-regime} & prices $z$ & non-dominant band & high-corn, high-soy & gross revenue & $1{,}600$ / $384$ \\
    \midrule
    \multicolumn{6}{@{}l}{\textit{Tier 2: single-axis schema shift}} \\
    \addlinespace[0.2em]
    \textbf{observation-schema} & observation $\Sigma_O$ & full schema $S_1$ & subset, rename, superset, anonymized & profit & $1{,}600$ / $1{,}024$ \\
    \textbf{action-menu} & action $\Sigma_A$ & A: \texttt{lnpkw}, \texttt{lnpk}; B: \texttt{lnpk}, \texttt{lw} & A: \texttt{lnw}, \texttt{ln}, \texttt{lw} (+ seen \texttt{lnpkw}); B: \texttt{lnpkw}, \texttt{lnw}, \texttt{ln} (+ seen \texttt{lw}) & profit & $2{\times}1{,}600$ / $512$ \\
    \textbf{reward-formulation} & objective $\Sigma_R$ & multi: yield, profit, water; ablation: yield only & yield, profit, water, nutrient & (varies) & $1{,}600{+}1{,}602$ / $512$ \\
    \midrule
    \multicolumn{6}{@{}l}{\textit{Tier 3: compositional schema and dynamics shift\quad(DSSAT, WOFOST, Cycles)}} \\
    \addlinespace[0.2em]
    \textbf{cross-simulator} & simulator: $\Sigma\!\times\!P$ & one source simulator & $3{\times}3$ source--target pairs & profit & $3{\times}1{,}600$ / $386$ \\
    \textbf{unified policy} & $\Sigma_O\!\times\!\Sigma_A\!\times\!\Sigma_R\!\times\!\text{Sim}$ & seen schema tuples & held-out schema combinations & (varies) & $3{,}200$ / $1{,}156$ \\
    \bottomrule
  \end{tabularx}
\end{table*}

\subsection{Baselines and metrics}
\label{sec:benchmark-protocol}

The target policy class is prompt-conditioned LLM-RL (Section~\ref{sec:framework-training}), evaluated in think and no-think variants. Figures show each arm's seed mean as a bar with $\pm$SD whiskers and mark the \emph{untrained model} (the same backbone before RL, scored under both response formats) on the same test scenarios as an open circle.

\textbf{NN-RL} (neural-network PPO): Stable-Baselines3 PPO~\citep{raffin2021sb3} over the numerical observation, with the interaction budget of LLM-RL and its one gradient pass per batch. This fixed-interface network applies to the Tier~1 settings (Section~\ref{sec:framework-pomdp}); potato weather-regime adds DQN and A2C, and the action-menu setting adds schema-capable PPO variants (Appendix~\ref{app:bl-nnppo}).

\textbf{Scripted agronomic reference}: a fixed rule set per simulator codified from agronomic guidance, such as crop-stage fertilizer and irrigation rules on WOFOST and a corn--soybean rotation for Cycles crop planning (Appendix~\ref{app:baselines}). It is uncalibrated against experts or optimal control.

\textbf{No-action}: the zero action at every decision, which applies no input on the crop-growth tasks (Appendix~\ref{app:bl-control}); falling below it is worse than abstaining.

\emph{Normalized return} places a policy relative to these two references:
\begin{equation}
  \label{eq:normalized-return}
  \tilde{R}(\pi) = \frac{R(\pi) - R_{\text{no-action}}}{\lvert R_{\text{ref}} - R_{\text{no-action}} \rvert},
\end{equation}
where $R(\pi)$ is the policy's mean test return and both references are scored on the same scenarios. No-action sits at $0$ and the scripted reference at $\pm 1$; $\tilde{R} > 1$ means the policy exceeds the reference. Where the two references are too close to divide by, as on weather-regime, we report the gain over no-action, $R(\pi) - R_{\text{no-action}}$. The gap varies across targets, so we compare policies by $\tilde{R}$ only within a target; Appendix~\ref{app:exp-normalization} lists raw returns, both reference returns and the gap for every headline cell.

\subsection{Experimental setup}
\label{sec:benchmark-setup}

LLM-RL trains Qwen3-4B-Instruct~\citep{yang2025qwen3} with GRPO~\citep{shao2024deepseekmath} on VeRL~\citep{sheng2024hybridflow}. The default reward is \texttt{profit\_max} (terminal storage-organ yield minus input costs); reward overrides and hyperparameters appear in Appendices~\ref{app:exp-per-setting} and~\ref{app:exp-protocol}. All main-text scores come from a frozen, prespecified held-out test suite of 6{,}278 evaluation rows, built after submission and disjoint from each experiment's training and validation data (Appendix~\ref{app:test-suite}). Each LLM-RL arm and NN baseline trains three seeds; we report mean $\pm$ SD over seeds, each run scored at its terminal checkpoint with no selection on validation, and give differences between arms in propagated SD (Appendix~\ref{app:exp-protocol}). These three-seed test-suite results are confirmatory; results only on validation or from a single seed (training curves, backbone comparisons) and the crop-variety per-crop breakdown are exploratory and labeled as such. We release the code, the test-suite manifests with generation configurations and seeds, disjointness audits, evaluation scripts, trained checkpoints, and per-run records for every reported table at \url{https://github.com/AgriManager-Framework/AgriManager} and \url{https://huggingface.co/AgriManager-Framework}, with a project page and documentation at \url{https://agrimanager-framework.github.io/} (licenses in Appendix~\ref{app:assets}).

\section{Results}
\label{sec:results}

\subsection{Tier 1: Within-schema distributional shift}
\label{sec:tier1}

\begin{figure*}[t]
  \centering
  \includegraphics[width=\linewidth]{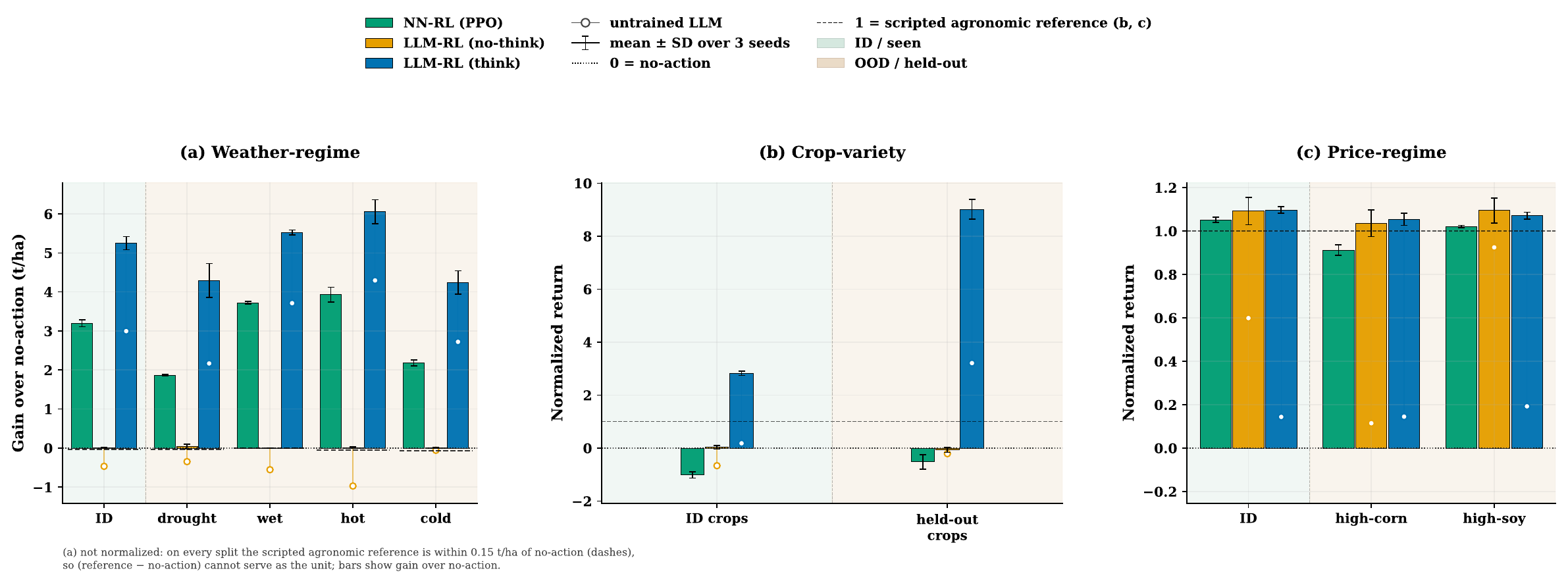}
  \caption{Tier 1 results, with bars and markers as in Section~\ref{sec:benchmark-protocol}; each panel has its own y-axis, with zero aligned. (a)~WOFOST weather-regime shift per split (chickpea and potato averaged per seed), as gain over no-action in profit-equivalent t/ha, since its two references are too close to normalize. (b)~WOFOST crop-variety shift on seen and held-out crops (4, 8, and 16 training crops averaged) and (c)~Cycles price-regime shift per regime, in normalized return.}
  \label{fig:tier1}
\end{figure*}

RL training lifts the think variant above the untrained model in every seed and split of Figure~\ref{fig:tier1} (30/30 seed--split pairs on weather-regime, 18/18 on crop-variety, 9/9 on price-regime), and every think seed scores above every seed of neural-network PPO. No-think tracks think on price-regime and stays near no-action on both WOFOST settings, where no-action is a stable basin of \texttt{profit\_max} and the no-think runs stop early under the rollout filter (Appendix~\ref{app:no-think-collapse}); the gap there reflects training stability.

\textbf{Weather-regime.}
Among the main-text methods, think scores highest on both crops in all five splits, the four out-of-distribution regimes included. It leads PPO by 1.16 to 3.54 profit-equivalent t/ha (3.6 to 15.5 propagated SD) on every split, and A2C, the strongest neural baseline on potato, by 4.5 to 7.3 propagated SD (Appendix~\ref{app:t1-neural}).

\textbf{Crop-variety.}
PPO averages below no-action on both seen and held-out crops at every coverage. In profit-equivalent t/ha, the unit of the per-crop table (Appendix~\ref{app:t12-percrop}), think leads it by 1.25 to 1.84 (5.6 to 32 propagated SD) on every coverage and split.
Per crop (exploratory), RL gains are uneven across the held-out crops, so subset-level heterogeneity appears even under a fixed schema.

\textbf{Price-regime.}
On multi-year Cycles crop planning (gross-revenue objective), think and no-think both reach 1.07, above PPO (0.99) and the untrained model (0.16 with thinking, 0.54 without). Their seed-matched difference is $0.00 \pm 0.06$: no observed advantage for reasoning. Per regime, PPO is the only trained arm below the reference (0.91 under high corn prices).

Within a fixed schema, RL training thus gives the think policy distributional generalization across weather, crop, and price shifts. Section~\ref{sec:tier2} releases $\Sigma$ and asks whether the same policy class transfers when the schema itself changes.

\subsection{Tier 2: Single-axis schema shift}
\label{sec:tier2}

\begin{figure*}[t]
  \centering
  \includegraphics[width=\linewidth]{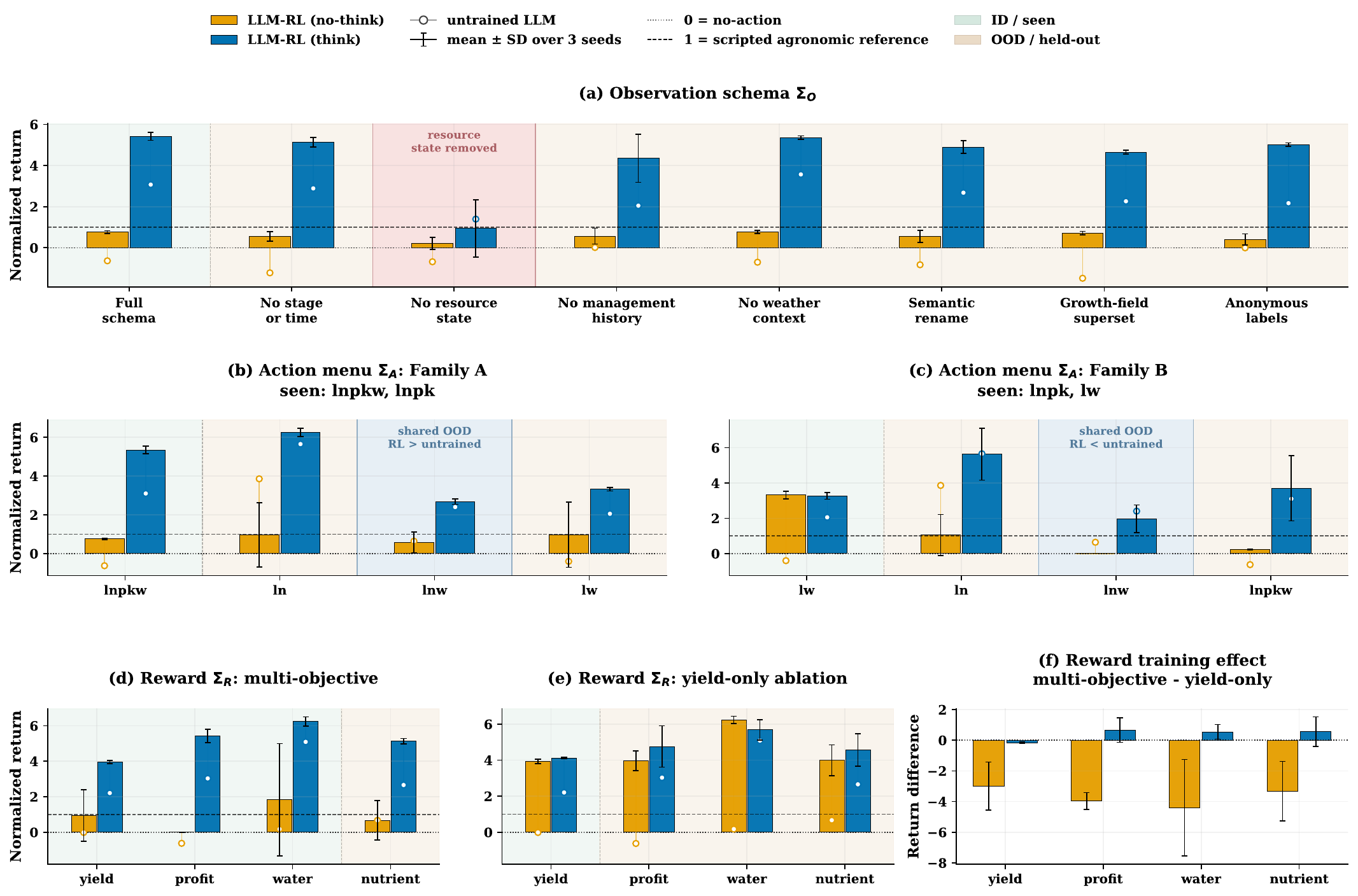}
  \caption{Tier 2 results in normalized return, with bars and markers as in Section~\ref{sec:benchmark-protocol}. (a)~Observation schema $\Sigma_O$, with the resource-state-removed variant highlighted. (b,c)~Action menu $\Sigma_A$, Families~A and~B, each family's seen menu first and the shared held-out menu \texttt{lnw} highlighted. (d,e)~Reward formulation $\Sigma_R$: multi-objective and yield-only training, with nutrient stewardship held out for both. (f)~Seed-matched difference per objective, multi-objective minus yield-only, for both variants.}
  \label{fig:tier2}
\end{figure*}

Figure~\ref{fig:tier2} reports Tier 2 results on the three schema axes.

\textbf{Observation-schema.}
On seven of the eight $\Sigma_O$ variants (the full schema, renamed or anonymized fields, a field superset, and drops of stage and time, management history, or weather context), think scores 4.4 to 5.4, and RL lifts it above the untrained model in every seed. Removing the resource state, which tells the policy whether an action is needed while unneeded actions still incur the input cost, drops think to $0.94 \pm 1.40$: the lowest variant in every seed and $0.45$ below the untrained model on average, with one of three seeds above it (Appendix~\ref{app:per-seed}). Observation-schema transfer is robust to representational change and depends on decision-critical state.

\textbf{Action-menu.}
Two training families share the held-out menu \texttt{lnw}. Family~A trains on \texttt{lnpkw} and \texttt{lnpk}, so it has seen N and water in one menu; Family~B trains on \texttt{lnpk} and \texttt{lw}, where they appear only separately. On \texttt{lnw}, think's RL gain over the untrained model is $+0.29$ for A and $-0.44$ for B. Family~A gains on all four test menus in every seed; Family~B gains in every seed only on its training menu \texttt{lw}. The sign flip is consistent with joint action coverage; each native menu also runs under its own production wrapper (Appendix~\ref{app:benchmark-t2}), which these runs do not separate from coverage. Schema-capable PPO baselines are reported in Appendix~\ref{app:bl-nnppo}.

\textbf{Reward-formulation.}
The backbone is trained on a multi-objective reward set (yield, profit, water stewardship) or on yield alone, and evaluated on four objectives with nutrient stewardship held out from both. Under think, both policies score above the untrained model on all four objectives in every seed, and the seed-matched difference (multi-objective minus yield-only) is small against scores of 3.9 to 6.2: $-0.17$ on yield, $+0.67$ on profit, $+0.55$ on water, and $+0.55$ on held-out nutrient. A think policy trained on yield alone thus transfers to three unseen objectives, and multi-objective training adds a modest margin. Under no-think, multi-objective training is unstable: two of its three seeds end at or near no-action on every objective, placing it below yield-only on all four.

The three axes therefore impose different burdens: $\Sigma_O$ requires information sufficiency, the $\Sigma_A$ results are consistent with joint action coverage, and on $\Sigma_R$, with reasoning, reward-schema transfer holds without objective coverage. Section~\ref{sec:tier3} shifts several axes at once.

\subsection{Tier 3: Compositional schema shift}
\label{sec:tier3}

\begin{figure*}[t]
  \centering
  \includegraphics[width=\linewidth]{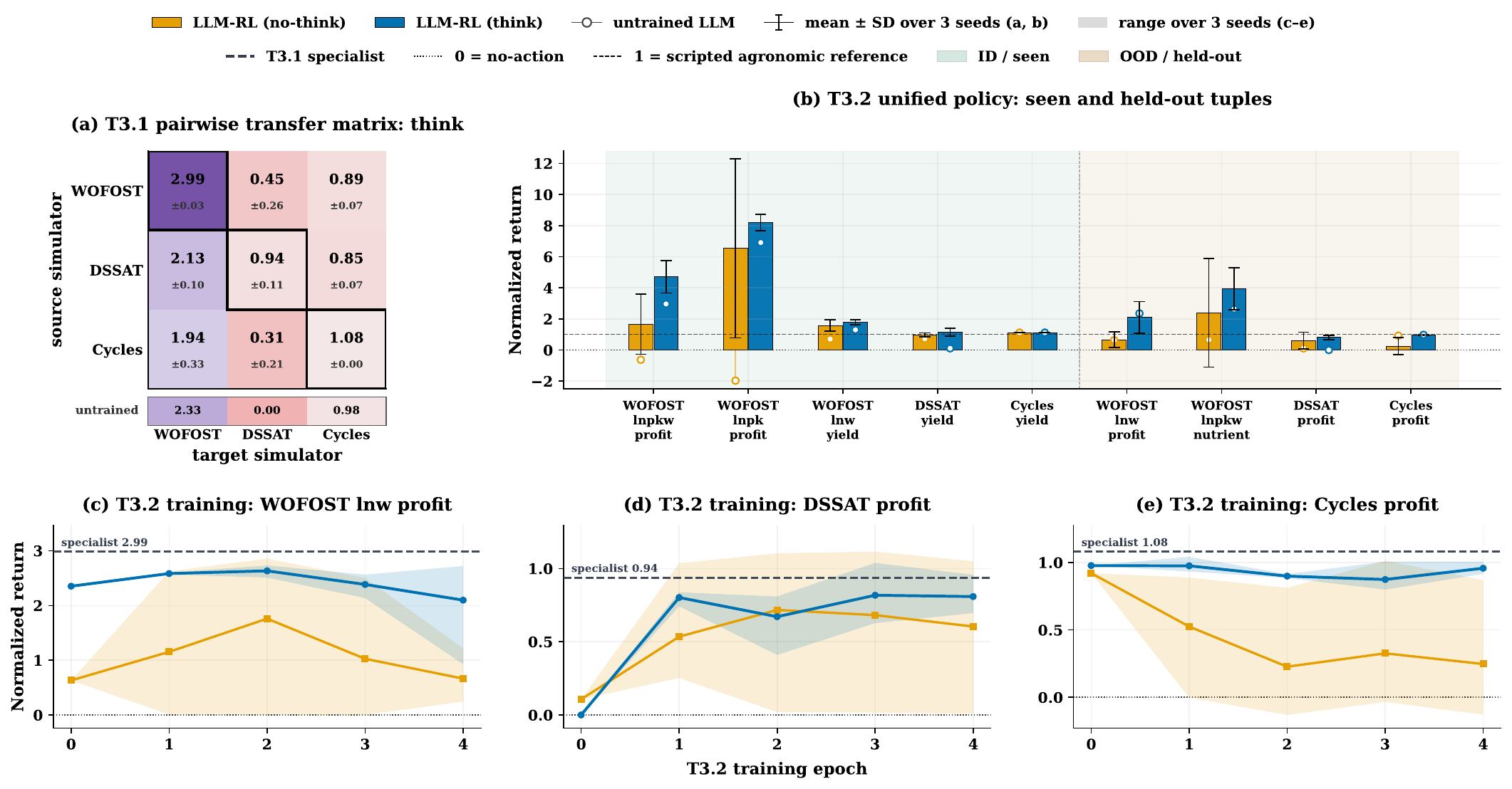}
  \caption{Tier 3 results in normalized return, with bars and markers as in Section~\ref{sec:benchmark-protocol}. (a)~T3.1 think-mode transfer matrix on maize profit; rows index the source simulator, columns the target, and cells compare within a column; the strip below gives the untrained model. (b)~T3.2 unified policy on five seen-control and four held-out tuples. (c--e)~Test-suite scores over T3.2 training (line: seed mean; band: seed range; intermediate points exploratory; epoch 0 is the untrained model) on the three held-out profit targets that call back T3.1; the dashed line gives the T3.1 specialist trained on that tuple.}
  \label{fig:tier3}
\end{figure*}

T3.1 switches simulator, which changes the transition dynamics $P$ together with the observation fields, the action menu and its units, the objective terms, and the site and weather source, so no single factor accounts for a transfer score (Appendix~\ref{app:t3-inventory}).

\textbf{Pairwise cross-simulator transfer.}
In Figure~\ref{fig:tier3}a, single-source training hurts four of the six transfers relative to the untrained model. The two into WOFOST (2.13 from DSSAT, 1.94 from Cycles, against 2.33) and the two into Cycles (0.89 from WOFOST, 0.85 from DSSAT, against 0.98) score below it in every seed. The transfers into DSSAT (0.45 from WOFOST, 0.31 from Cycles) gain only relative to an untrained think model that fails the output format on every DSSAT decision and therefore scores as no-action (0.00; Appendix~\ref{app:t3-details}). DSSAT is the hardest target: transfers into it recover 33\% to 48\% of the DSSAT-trained policy's gain over no-action, against 65\% to 71\% into WOFOST and 79\% to 82\% into Cycles, and part of the shortfall is format failure. The pattern is consistent with a limit set by the transition dynamics $P$, which the prompt never describes and these runs do not isolate.

\textbf{Toward a unified prompt-conditioned policy.}
T3.2 trains a single backbone on schema tuples of the form (simulator, action menu, objective), with four target tuples held out; every component of every held-out tuple appears in at least one training tuple (Appendix~\ref{app:benchmark-t3}). The think unified policy scores above no-action on all four held-out tuples in every seed. It gains over the untrained model on WOFOST \texttt{lnpkw} nutrient and on DSSAT profit, partly by learning the output format there. On DSSAT it reaches $0.81 \pm 0.13$ against $0.94 \pm 0.11$ for the matched T3.1 specialist, which trains on this exact tuple and is scored on the same scenarios, about 86\% of the specialist's gain, with overlapping seeds; the gain arrives in the first training epoch (Figure~\ref{fig:tier3}d). On WOFOST \texttt{lnw} profit and Cycles profit it ends below the untrained model (2.10 and 0.96 against 2.35 and 0.98), and every specialist seed scores above every unified seed (specialists 2.99 and 1.08; Appendix~\ref{app:t3-details}); the wide WOFOST spread comes from one seed.

Without thinking, the unified policy is unstable across seeds: no-think ends below think on all four held-out targets, and one of its three seeds ends near or below no-action on each of them. This matches the no-think instability on WOFOST in Tier~1 and under multi-objective rewards in Tier~2 (Sections~\ref{sec:tier1} and~\ref{sec:tier2}). The association of thinking with stable training holds for the Qwen backbones we tested and reverses for Llama-3.2-3B, where a single-seed unified pair has no-think ahead on four of the five WOFOST tuples (Appendix~\ref{app:t3-backbones}).

Tier 3 therefore gives the thesis question a partial answer. Trained only on other combinations of the same components, one prompt-conditioned policy recovers about 86\% of the in-distribution specialist's gain on DSSAT and stays below it on the WOFOST and Cycles profit targets, while single-source training mostly hurts transfer. Closing the gap between recombining schema components and training on the target tuple itself remains open.

\section{Discussion}
\label{sec:discussion}

Generalization in agricultural RL depends on which part of the interface shifts, so a single generalization score can hide the axis that fails. Our settings suggest three lessons for building such policies (Appendix~\ref{app:cross-tier} summarizes each tier). Observations can be renamed or extended, but the state that decides whether an input is needed must stay in the schema. Transfer to a new action menu is consistent with having seen its actions together in one training menu. With reasoning, a policy trained on one objective can be redirected to new objectives through the prompt. On Qwen, reasoning acts mainly as a training stabilizer: without it, RL often settles near no-action when inputs carry a cost, and a single-seed Llama-3.2-3B pair does not show this effect (Appendix~\ref{app:t3-backbones}). The open problem is compositional shift across simulators, where recombining seen components recovers only part of a target-trained specialist's gain; changes in transition dynamics, which the prompt does not describe, are one candidate bottleneck; describing dynamics in the prompt or adapting at test time are natural next steps.

\paragraph{Limitations.}
All evaluation is in simulation; on-farm validation is the next step. Three seeds per arm give direction and spread without formal significance tests, and normalized return measures distance from codified practice with no optimality bound (Appendix~\ref{app:discussion-details}).

\paragraph{Broader impacts.}
The benchmark lets researchers test agricultural policies before any deployment. Its main risk is over-reading simulator results as farm advice: released policies should not schedule inputs or planting without local agronomic review and field validation.

\begin{ack}
This work was supported by the AIFARMS National AI Institute (USDA NIFA Award Number 2020-67021-32799) and the Center for Digital Agriculture at the University of Illinois. The views expressed and conclusions drawn herein are solely those of the authors and do not necessarily reflect the official policies or positions, expressed or implied, of any sponsoring organizations. This work used the Delta advanced computing and data resource at the University of Illinois Urbana-Champaign and its National Center for Supercomputing Applications through allocations CIS251101, CIS260670, and CIS260693 from the Advanced Cyberinfrastructure Coordination Ecosystem: Services \& Support (ACCESS) program, which is supported by U.S. National Science Foundation grants \#2138259, \#2138286, \#2138307, \#2137603, and \#2138296. Jeremiah Barr was employed by Corteva Agriscience when this work was done.
\end{ack}

\bibliographystyle{plainnat}
\bibliography{references}

\appendix

\clearpage
\etocdepthtag.toc{mtappendix}
\etocsettagdepth{mtmain}{none}
\etocsettagdepth{mtappendix}{subsection}
\etocsettagdepth{mtchecklist}{none}
\etocsettocstyle{\section*{Appendix contents}}{}
\tableofcontents
\clearpage

\section{Framework details}
\label{app:adapter-contract}

A new simulator joins the \AgriManager{} framework by adding one package under \texttt{agrimanager/env/<name>/} that implements a documented adapter; the framework discovers it by folder name, with no central registry or routing edit. Each environment is exposed through three interfaces: a dataset generator that materializes scenarios into a parquet file, a multi-turn environment for LLM rollout, and a numeric adapter that bridges the same parquet to a Gymnasium environment for the fixed-interface NN-PPO baseline (Appendix~\ref{app:bl-nnppo}). The parquet artifact is the seam between dataset build and training: every row carries the full environment configuration, so the two phases stay decoupled and the same split always reconstructs the same environment. The released code documents the full method-level contract.

\subsection{Simulator pool}
\label{app:simulator-pool}

\textbf{DSSAT} (Decision Support System for Agrotechnology Transfer): a process-based crop-growth simulator suite from the IBSNAT/USDA line~\citep{jones2003dssat}. Our adapter (\texttt{gym\_dssat}) builds on gym-DSSAT~\citep{gautron2022gymdssat}, defaults to maize and adds prompt configurations for cotton and rice; it supports the LLM path.

\textbf{WOFOST} (WOrld FOod STudies): the Wageningen crop-growth simulator behind the EU MARS yield-forecasting system~\citep{dewit2019wofost}, run through its Python implementation PCSE. Our adapter (\texttt{wofost\_gym}) builds on WOFOSTGym~\citep{solow2025wofostgym} and provides trait files for over twenty crops, six action-menu configurations (the full \texttt{lnpkw} menu of N/P/K and water plus five restricted subsets), and four management objectives (\texttt{yield\_max}, \texttt{profit\_max}, \texttt{water\_stewardship}, \texttt{nutrient\_stewardship}); single-axis schema variants along $\Sigma_O$, $\Sigma_A$, and $\Sigma_R$ can all be constructed on this backend. WOFOST is the reference instantiation and supports both the LLM and NN paths.

\textbf{Cycles}: the Penn State multi-year agroecosystem model~\citep{kemanian2024cycles}, the only simulator in the pool that contributes a crop-planning task, a 19-year \texttt{corn--soybean} rotation in which annual crop choices drive long-horizon soil carbon and nitrogen dynamics; it also provides a single-season maize nitrogen-fertilization task. The adapter (\texttt{cycles\_gym}) builds on CyclesGym~\citep{turchetta2022cyclesgym}; it supports the LLM path for both tasks and the NN path for crop planning. Appendix~\ref{app:assets} lists the version and license of each simulator, wrapper, model, and software package.

\FloatBarrier
\section{Benchmark task specification}
\label{app:benchmark-details}
\label{app:exp-per-setting}

This appendix specifies each of the eight benchmark settings of
Table~\ref{tab:test-sets-overview}: what it varies, its agricultural
interpretation, its training, validation and test scenarios, and its objective.
Table~\ref{tab:exp-splits} collects the split sizes; the per-setting paragraphs
below give the construction details. Throughout, $\Sigma=(\Sigma_O,\Sigma_A,\Sigma_R)$
is the policy-facing schema of Section~\ref{sec:framework-pomdp}.

\begin{table*}[t]
  \centering
  \footnotesize
  \setlength{\tabcolsep}{3.5pt}
  \renewcommand{\arraystretch}{1.22}
  \caption{Per-setting training, validation and test split sizes. ``$n \times m$'' denotes $n$ groups of $m$ scenarios; the paragraphs below define the groups. Validation sets serve training-time monitoring only; every main-text score uses the test suite (Appendix~\ref{app:test-suite}). The three test groups of 130 are the rebuilt Cycles suites.}
  \label{tab:exp-splits}
  \begin{tabularx}{\textwidth}{@{}
                  >{\hsize=1.0\hsize}Y
                  >{\hsize=.48\hsize}Y
                  >{\hsize=.95\hsize}Y
                  >{\hsize=1.27\hsize}Y
                  >{\hsize=1.30\hsize}Y@{}}
    \toprule
    Setting & Simulator & Train & Validation (named groups) & Test (named groups) \\
    \midrule
    T1.1 weather-regime         & WOFOST    & 1600 per crop                                 & $5{\times}128$ per crop: id, drought, wet, hot, cold & $5{\times}128$ per crop: id, drought, wet, hot, cold \\
    T1.2 cross-crop (4ID)       & WOFOST    & $4{\times}400$                                & $4{\times}96$ ID + $4{\times}96$ OOD (= 768) & $4{\times}128$ seen + $4{\times}128$ held-out (= 1024), shared by 4ID/8ID/16ID \\
    T1.2 cross-crop (8ID)       & WOFOST    & $8{\times}200$                                & $8{\times}48$ ID + $4{\times}96$ OOD (= 768) & shared T1.2 suite \\
    T1.2 cross-crop (16ID)      & WOFOST    & $16{\times}100$                               & $16{\times}24$ ID + $4{\times}96$ OOD (= 768) & shared T1.2 suite \\
    T1.3 price-regime           & Cycles    & 1600                                          & $3{\times}128$ paired: id\_test, high\_corn, high\_soybean & $3{\times}128$ paired, same regimes \\
    T2.1 observation-schema     & WOFOST    & 1600                                          & $8{\times}128$ shared scenarios: S1, S2a--d, S3, S4, S5 & $8{\times}128$ shared scenarios, same variants \\
    T2.2 action-menu (Family A) & WOFOST    & $2{\times}800$: lnpkw, lnpk                   & $2{\times}128$ seen + $3{\times}128$ held-out: lnw, ln, lw & $4{\times}128$ paired: lnpkw, lnw, ln, lw \\
    T2.2 action-menu (Family B) & WOFOST    & $2{\times}800$: lnpk, lw                      & $2{\times}128$ seen + $3{\times}128$ held-out: lnpkw, lnw, ln & same suite as Family~A \\
    T2.3 reward-formulation     & WOFOST    & 1600: yield, profit, water (534/533/533); yield-only: 534 scenarios $\times3$ (1602 rows) & $4{\times}128$: yield, profit, water, nutrient & $4{\times}128$ paired, same objectives \\
    T3.1 cross-simulator        & W / D / C & 1600 per source                               & $3{\times}128$: WOFOST, DSSAT, Cycles target & 128 WOFOST + 128 DSSAT + 130 Cycles \\
    T3.2 unified policy         & all three & $8{\times}400$ (= 3200)                       & $5{\times}128$ seen-control + $4{\times}128$ held-out & $7{\times}128$ + $2{\times}130$ (Cycles profit, yield) \\
    \bottomrule
  \end{tabularx}
\end{table*}

\FloatBarrier
\subsection{Frozen test suite}
\label{app:test-suite}

Every main-text score comes from a test suite built after submission. Its suites were frozen on 25--27 July 2026, during the author-response period, and the three Cycles crop-growth suites were rebuilt on 26 September 2026 (below). The numerical renderings of the T1.2 and T1.3 suites that the neural baselines read were added on 3 September 2026, scenario for scenario identical to the frozen suites. The suite holds 6{,}278 evaluation rows over the eight settings in groups of 128 (130 for the three rebuilt Cycles suites): 2{,}688 for Tier~1, 2{,}048 for Tier~2 and 1{,}542 for Tier~3 (Table~\ref{tab:exp-splits}). Rows are not all distinct physical scenarios, by design: T2.1 renders one set of 128 scenarios under eight observation schemas, the Tier~2 suites, the WOFOST target of T3.1 and the WOFOST tuples of T3.2 share one base of 128 maize scenarios (Appendix~\ref{app:benchmark-t2}), T1.3 renders each of 128 scenarios under three price vectors, the DSSAT target of T3.1 and the DSSAT profit tuple of T3.2 use the same generator seeds, and each rebuilt Cycles suite repeats 13 weather years ten times. Each suite is fixed by a generation configuration and seed, needs no checkpoint to build, and comes with a manifest, checksums and an audit of its overlap with the training and validation scenarios of the matching experiment. The audit compares physical scenario keys: crop, year, latitude and longitude on the WOFOST pools, the generator seed on DSSAT, and the weather year on Cycles. Policies, think or no-think, are scored on the same suite under the response format they were trained with.

The three Tier~3 Cycles crop-growth suites (the T3.1 target and the T3.2 profit and yield tuples) were rebuilt. A Cycles crop-growth scenario is fixed by its weather year at a single site (Rock Springs); the first build was audited by scenario identifier and reused the eight validation years (1996--2003). The rebuilt suites draw on weather years 2004--2016, which training (1980--1995) and validation leave unused. Their effective sample is 13 physical scenarios, so scenario-level standard errors on these three targets overstate precision; seed-level SDs are unaffected. Every Cycles crop-growth number in this version uses the rebuilt suites, except the exploratory Qwen3-8B format diagnostic of Appendix~\ref{app:t3-backbones}, scored before the rebuild.

These three-seed test-suite results at the terminal checkpoint are confirmatory. Results available only on validation or from a single training seed are exploratory: the training curves (Appendix~\ref{app:full-results-training-curves}), the single-seed PPO~\texttt{n8} cells on T1.1, and the backbone comparisons (Appendix~\ref{app:t3-backbones}). The Tier~1 lower-tail readout (Appendix~\ref{app:t1-neural}) is also exploratory: it uses all three seeds but a later re-scoring pass and is not a prespecified target. The T1.2 per-crop breakdown (Appendix~\ref{app:t12-percrop}) is also exploratory: it uses the three-seed test suite and splits it below the prespecified targets. The Tier~3 training curves of Figure~\ref{fig:tier3}(c--e) score intermediate checkpoints of the three seeds on the test suite; the protocol fixes the terminal checkpoint, so their intermediate points are exploratory and their end points equal the confirmatory bars of panel~(b).

\FloatBarrier
\subsection{Tier 1: within-schema distributional shifts}
\label{app:benchmark-t1}

Tier~1 keeps the observation, action, and reward schema fixed and changes only
the episode context, testing whether a policy that needs no interface change is
still robust to realistic agricultural variation.

\textbf{T1.1 Weather-regime shift.} Two single-crop runs (chickpea, potato)
train on ordinary viable weather and are evaluated on four named regimes plus an
in-distribution control, testing whether fertilizer and irrigation timing track
phenology and water stress. Validation regimes are defined on WOFOST
growing-season weather statistics with $p{=}0.20$: \texttt{drought}/\texttt{wet}
are the lowest/highest $p$ fraction of cumulative rainfall, \texttt{cold}/\texttt{hot}
the lowest/highest $p$ of mean temperature; each contributes 128 scenarios. The test
regimes are rebuilt from the 1{,}600 pool scenarios per crop that training and
validation leave unused, under a predeclared rule computed over each crop's
sowing-to-maturity window: each group takes one fifth of these scenarios (320),
assigned in turn: the 320 coldest by mean temperature form \texttt{cold}, the
320 hottest remaining scenarios \texttt{hot}, the 320 driest remaining by
cumulative rainfall \texttt{drought} and the 320 wettest remaining \texttt{wet},
and the last 320 form \texttt{id}; 128 scenarios are drawn from each group with a fixed seed. The test \texttt{id} group is
therefore more moderate than a random in-distribution draw.

\textbf{T1.2 Cross-crop trait shift.} Crop identity and traits change while the
WOFOST observation fields, N/P/K/water actions, and \texttt{profit\_max} reward
stay fixed; one interface represents every crop, so this is a context shift. Traits enter the prompt as a 23-dimensional
descriptor (temperature adaptation, phenology, rooting and water response,
nutrient capacity, assimilation, storage allocation). Three coverage settings
(4ID/8ID/16ID) hold a fixed 1600-scenario train budget while varying the number
of training crops ($4{\times}400$, $8{\times}200$, $16{\times}100$); all three
are evaluated on the same four held-out crops (\texttt{barley},
\texttt{groundnut}, \texttt{pigeonpea}, \texttt{seed\_onion}) and total 768
validation rows. The test suite draws 128 scenarios per crop from the crop pool's
own test split for the four held-out crops and for the four 4ID training crops
(\texttt{cotton}, \texttt{millet}, \texttt{rapeseed}, \texttt{rice}), which the
8ID and 16ID settings also train on.

\textbf{T1.3 Price-regime shift.} A 19-year crop-planning episode on Cycles
(crops \texttt{CornRM.100}, \texttt{SoybeanMG.3}) keeps the planting action and
gross-revenue objective fixed and changes only the crop-price vector. Gross
revenue is yield times price and carries no input-cost term; its formula stays
fixed while its parameters move, so this is an economic-context shift within one
reward formulation. Training samples non-dominant in-distribution price pairs
matching a Corn Belt rotation. Validation is paired: identical weather, location,
and seed produce three rows differing only in the price vector, one per regime in
\{\texttt{id\_test}, \texttt{high\_corn}, \texttt{high\_soybean}\}, for
$128{\times}3=384$ rows. The test suite repeats this pairing on 128 new scenarios
from a fresh seed range.

\FloatBarrier
\subsection{Tier 2: single-axis schema shifts}
\label{app:benchmark-t2}

Tier~2 targets one component of $\Sigma$ between training and test; on the action axis each menu also brings its own production wrapper (T2.2 below).
The target schema is supplied in the prompt, while the simulator stays WOFOST
maize. The Tier~2 test suites, the WOFOST target of T3.1 and the WOFOST tuples of
T3.2 render one frozen base of 128 maize scenarios, disjoint from the submitted
training and validation scenarios.

\textbf{T2.1 Observation-schema shift ($\Sigma_O$).} Training uses the full
15-field WOFOST reference schema \texttt{S1} (development stage, storage-organ
weight, nutrients, soil moisture, cumulative inputs, weather, day count).
Validation re-renders one shared set of 128 maize scenarios under eight
projections of the same underlying state: \texttt{S1} (in-distribution
reference); four missing-information masks \texttt{S2a}--\texttt{S2d} that drop
the stage/time, resource-state, management-history, or weather-context fields;
\texttt{S3} renaming labels to agronomic synonyms; \texttt{S4} adding seven
growth and uptake fields; and \texttt{S5} replacing labels with anonymous
\texttt{A}--\texttt{O} plus a glossary. The prompt is therefore responsible for
telling the policy what each field means; the $8{\times}128=1024$ rows are paired
across schemas, in validation and in the test suite alike.

\textbf{T2.2 Action-menu shift ($\Sigma_A$).} Prompts expose native WOFOST menus
that model different equipment or input-access constraints: \texttt{lnpkw}
(N/P/K/water), \texttt{lnpk}, \texttt{lnw} (N/water), \texttt{ln}, and
\texttt{lw}. Two families with matched budgets train on two menus each (Family~A:
\texttt{lnpkw}, \texttt{lnpk}; Family~B: \texttt{lnpk}, \texttt{lw}; 800
scenarios per menu) and are validated on all five (128 paired maize scenarios
each), so the held-out menus are \texttt{lnw}/\texttt{ln}/\texttt{lw} for A and
\texttt{lnpkw}/\texttt{lnw}/\texttt{ln} for B. The test suite covers the four
reported menus (\texttt{lnpkw}, \texttt{lnw}, \texttt{ln}, \texttt{lw}). The
reward is \texttt{profit\_max} throughout, with input costs absorbed into the
objective; success requires reallocating the remaining controls when an input is
unavailable. Each native menu runs under its own WOFOST production-level wrapper,
in which the inputs absent from the menu are simulated as non-limiting
(\texttt{lw}, for example, is water-limited with unlimited nutrients). A menu
change therefore also changes the crop's resource limits: on the same test
scenarios, no-action returns 0.12 on \texttt{lnw}, 0.15 on \texttt{ln}, 0.40 on
\texttt{lnpkw} and 2.77 on \texttt{lw}. The T2.2 runs do not separate action
coverage from this wrapper.

\textbf{T2.3 Reward-formulation shift ($\Sigma_R$).} Observation and actions stay
fixed while the prompt objective and terminal reward change among
\texttt{yield\_max}, \texttt{profit\_max}, \texttt{water\_stewardship}, and
held-out \texttt{nutrient\_stewardship}. With $Y$ the final storage-organ weight,
$y=Y/Y_{\mathrm{ref}}$, and $h(z)=\max(0,z)^2$,
\[
\begin{aligned}
R_{\mathrm{yield}} &= y,\\
R_{\mathrm{profit}} &= (Y-3.5N-5.5P-2.25K-0.05W)/Y_{\mathrm{ref}},\\
R_{\mathrm{water}} &= y-0.10(W/500)-4h(0.80-y)-2h(W/500-1).
\end{aligned}
\]
Nutrient stewardship reuses the yield floor but penalizes N/P/K over-application
and terminal nutrient pools outside a target range, so the held-out objective is
related to but distinct from the seen ones. Training mixes the three
in-distribution objectives and is validated and tested on all four; a per-scenario
$Y_{\mathrm{ref}}$ calibrated by a scripted probe places the four rewards on a
common scale. A yield-only ablation trains on \texttt{yield\_max} alone (and
validates on the same four objectives) to isolate the effect of training-time
reward diversity.

\FloatBarrier
\subsection{Tier 3: compositional schema and simulator shifts}
\label{app:benchmark-t3}

Tier~3 changes multiple schema components together; T3.1 additionally changes
simulator dynamics $P$.

\textbf{T3.1 Cross-simulator maize transfer.} Three single-source specialists
(one per simulator, 1600 train and 128 validation scenarios each) all optimize
grain-equivalent maize profit over the inputs the target simulator exposes,
\[
\mathrm{profit}_{\mathrm{ge}}=\mathrm{yield}_{\mathrm{kg/ha}}
-3.5\,\mathrm{totalN}_{\mathrm{kg/ha}}
-0.05\,\mathrm{totalIrrig}_{\mathrm{mm}},
\]
omitting the water term where irrigation is unavailable. The objective text is
shared, but prompts stay simulator-native: WOFOST exposes storage-organ and
development state with discrete N/water actions, DSSAT exposes crop--soil
variables with bounded N/irrigation actions, and Cycles exposes
soil-nitrogen/crop-growth state with nitrogen management only. Each specialist is
evaluated on all three targets, forming the $3\times3$ transfer matrix whose
diagonal is the within-simulator reference and whose off-diagonal cells are
zero-shot cross-simulator transfer.

\textbf{T3.2 Unified prompt-conditioned policy.} A single policy is trained
jointly on eight schema tuples spanning three simulators, three action menus, and
four objectives (400 rows per tuple, 3200 total), following a marginal-coverage
design with exact-combination holdout: every held-out simulator, menu, and
objective appears in some training tuple, but the deployment tuple itself does
not, which separates memorizing whole task cards from recombining schema factors.
The four held-out validation tuples (WOFOST \texttt{lnw/profit}, WOFOST
\texttt{lnpkw/nutrient}, DSSAT \texttt{profit}, Cycles \texttt{profit}) combine the
Tier~2 action or reward axes with the Tier~3 simulator axis; five seen-control
tuples provide an in-distribution reference for the same trained policy. All
validation sets use 128 rows. The WOFOST tuples keep the submitted T3.2
calibration, a fixed reference yield $Y_{\mathrm{ref}}=8500$\,kg/ha, so their
scores are on a different scale from the same objectives in T2.3.

\paragraph{What changes in each cross-simulator transfer.}
\label{app:t3-inventory}
Every T3.1 transfer keeps the crop (maize), the prompt template, and the form of the profit objective, and changes the transition dynamics $P$ together with several schema and scenario factors (Table~\ref{tab:t3-inventory}). The three simulators differ as follows. WOFOST (\texttt{lnw} menu) offers nitrogen or irrigation at four discrete levels each, over 24 decisions at 10-day intervals, on maize scenarios drawn from the maize weather pool (3{,}840 site-years at 3{,}838 locations between 39 and 49\,$^\circ$N and 10\,$^\circ$W and 30\,$^\circ$E, weather years 1984--2019, NASA POWER daily weather); its observation lists development stage, storage-organ mass, soil nutrient and moisture state, cumulative inputs, and daily weather. DSSAT offers free amounts of nitrogen (below 30\,kg/ha) and irrigation (below 40\,mm) per decision, alone or together, over 24 decisions at 10-day intervals, with scenarios that differ by generator seed at a single site (climate file UFGA, the gym-DSSAT maize default) and a fixed soil profile; its observation lists days after planting, growth stage, grain and aboveground biomass, leaf area index, nitrogen and water stress, nitrogen uptake, daily weather, and cumulative inputs. Cycles offers nitrogen only, at eleven levels from 0 to 150\,kg/ha, over 26 decisions at 10-day intervals from 1 January, at a single site (Rock Springs) where a scenario is fixed by its weather year; its observation lists 27 fields: three site constants (latitude, altitude, screening height), the day's weather (precipitation, maximum and minimum temperature, solar radiation, maximum and minimum relative humidity, wind), crop stage and thermal time, cumulative, aboveground and root biomass, fraction of intercepted radiation, total, aboveground and root nitrogen, aboveground nitrogen concentration, nitrogen fixation, nitrogen added, nitrogen and water stress, potential transpiration, day of year, and cumulative applied nitrogen. Grain yield is not among them. Where irrigation is unavailable (Cycles) the objective drops its water term, and each target is normalized by its own scripted reference. Transfer scores are therefore compared only within a target.

\begin{table}[t]
  \centering
  \footnotesize
  \setlength{\tabcolsep}{3pt}
  \renewcommand{\arraystretch}{1.15}
  \caption{What changes in each T3.1 transfer, source $\to$ target. Every transfer also changes the transition dynamics $P$ and the observation fields (simulator-native names, units, and grouping), and keeps the crop (maize) and the profit form $\text{yield}-3.5\,N-0.05\,\text{irrigation}$.}
  \label{tab:t3-inventory}
  \begin{tabularx}{\linewidth}{@{}l Y Y Y c@{}}
    \toprule
    Transfer & Action menu & Objective terms & Site and weather & Decisions \\
    \midrule
    WOFOST $\to$ DSSAT   & discrete levels $\to$ free amounts; N and water may combine & unchanged & weather pool $\to$ one DSSAT site & $24\to24$ \\
    WOFOST $\to$ Cycles & water removed; N levels $4\to11$ & water term dropped & weather pool $\to$ Rock Springs years & $24\to26$ \\
    DSSAT $\to$ WOFOST   & free amounts $\to$ discrete levels; one input per decision & unchanged & one DSSAT site $\to$ weather pool & $24\to24$ \\
    DSSAT $\to$ Cycles & water removed; free amounts $\to$ 11 N levels & water term dropped & one DSSAT site $\to$ Rock Springs years & $24\to26$ \\
    Cycles $\to$ WOFOST & water added; N levels $11\to4$ & water term added & Rock Springs years $\to$ weather pool & $26\to24$ \\
    Cycles $\to$ DSSAT & water added; 11 N levels $\to$ free amounts & water term added & Rock Springs years $\to$ one DSSAT site & $26\to24$ \\
    \bottomrule
  \end{tabularx}
\end{table}

\FloatBarrier
\section{Prompt templates}
\label{app:prompts}

This appendix gives the prompt contract used by the LLM policies and identifies which blocks change across benchmark settings. The user prompt is rebuilt at each decision step around three schema-carrying blocks, one per axis of $\Sigma$.

\FloatBarrier
\subsection{Schema-carrying user prompt}
\label{app:prompts-user}

\begin{figure}[H]
  \centering
  \begingroup
  \definecolor{schemaReward}{HTML}{4A90D9}
  \definecolor{schemaObserve}{HTML}{5BA58B}
  \definecolor{schemaAction}{HTML}{D4A252}
  \setlength{\fboxrule}{0.5pt}%
  \newcommand{\schemaBlock}[5]{%
    \par\noindent
    \begingroup
    \setlength{\fboxsep}{0pt}%
    \fcolorbox{#1!60!black}{#1!7}{%
      \begin{minipage}{\dimexpr\linewidth-2\fboxrule\relax}%
        \parindent=0pt\relax
        \colorbox{#1!22}{%
          \makebox[\linewidth][l]{%
            \rule[-0.5em]{0pt}{1.55em}%
            \hspace*{0.85em}%
            \sffamily\small\bfseries\textcolor{#1!48!black}{$#2$\hspace*{0.55em}\MakeUppercase{#3}}%
            \hfill
            \normalfont\sffamily\footnotesize\textcolor{#1!48!black}{#4}%
            \hspace*{0.85em}%
          }%
        }%
        \par\nointerlineskip
        {\color{#1!50!black}\hrule height 0.4pt}%
        \vskip 0.55em
        \begingroup
          \leftskip 0.85em\relax
          \rightskip 0.85em plus 1fill\relax
          \ttfamily\footnotesize
          #5\par
        \endgroup
        \vskip 0.5em
      \end{minipage}%
    }%
    \endgroup
    \par\vskip 0.7em
  }
  \begin{minipage}{0.98\linewidth}
  \schemaBlock{schemaReward}{\Sigma_R}{reward schema}
    {\texttt{\textless management objective\textgreater} block}
    {\textless management objective id="profit\_max"\textgreater\\
     \hspace*{1.25em}profit = final\_WSO\\
     \hspace*{1.25em}\phantom{profit }{} - 3.5*N - 5.5*P\\
     \hspace*{1.25em}\phantom{profit }{} - 2.25*K - 0.5*irrig\\
     \textless/management objective\textgreater}

  \schemaBlock{schemaObserve}{\Sigma_O}{observation schema}
    {\texttt{\textless current observation\textgreater} block}
    {\textless current observation\textgreater\\
     \hspace*{1.25em}crop:\hspace*{2.0em}DVS=-0.099,\ WSO=0 kg/ha\\
     \hspace*{1.25em}soil:\hspace*{2.25em}N=2.61,\ P=2.61,\ K=2.61\\
     \hspace*{1.25em}weather:\hspace*{1.1em}rain=1.5 cm,\ T=6.84 C\\
     \textless/current observation\textgreater}

  \schemaBlock{schemaAction}{\Sigma_A}{action schema}
    {Available actions block}
    {Available actions (pick exactly one):\\
     \hspace*{1.25em}Do nothing\\
     \hspace*{1.25em}Apply N/P/K fertilizer: 20, 40, 60, 80 kg/ha\\
     \hspace*{1.25em}Irrigate: 5, 10, 15, 20 cm}
  \end{minipage}
  \endgroup
  \caption{Abridged user prompt from a chickpea episode. The three boxed blocks carry the reward, observation, and action axes of $\Sigma=(\Sigma_O,\Sigma_A,\Sigma_R)$ defined in Section~\ref{sec:framework-pomdp}; the full prompt also includes system instructions, episode context, interaction history, and the response-format contract. Prompts expose field names, units, action arguments, and the objective formula, but not hidden simulator state, future weather, or transition parameters.}
  \label{fig:prompt-template}
\end{figure}

\FloatBarrier
\subsection{System prompt and response-format variants}
\label{app:prompts-response}

All LLM environments use a short role-setting system prompt that keeps the task semantics out of the system message (they are supplied by the episode-level user prompt, so schema changes are controlled through dataset fields):

\begin{verbatim}
You are an agricultural management expert. Optimize the management objective
specified in the user message. Use only the available actions and respond in
the required format.
\end{verbatim}

We evaluate two response contracts. In the \textbf{think} variant, the user prompt asks the model to reason briefly before acting and to answer in a fixed format (abridged; the full wording is in the listings of Appendix~\ref{app:prompts-full}):

\begin{verbatim}
Please reason briefly before answering, then respond using the exact format:
<tool_call> brief reasoning </tool_call> <answer> final action </answer>
\end{verbatim}

In the \textbf{no-think} variant the same state and schema are shown, but the contract removes the reasoning block:

\begin{verbatim}
Respond using the exact format: <answer> ... </answer> with no extra text.
\end{verbatim}

The parser executes only the action inside the \texttt{\textless answer\textgreater} block. The environment adapter maps malformed responses and actions outside the declared menu to a fallback action, and the prompt stays unchanged (Appendix~\ref{app:invalid-rates} reports how often this happens).

\FloatBarrier
\subsection{Complete prompts for the two task families}
\label{app:prompts-full}

Each benchmark setting is a schema delta on one of two task families (Table~\ref{tab:prompt-deltas}): a within-season \emph{crop-growth} task that sets fertilization and irrigation at every step, and a multi-year \emph{crop-planning} task that picks the crop and planting week once per year. Both families are multi-turn episodes: a season or rotation runs for many decision steps, and the policy is queried once per step with the current state and the interaction history so far. The two families differ in system role, observation fields, action format, and decision horizon. Each listing below is a single such turn, from the system message to the model response, exactly as the policy receives and emits it; the turn index in the \texttt{=== USER ===} header gives its position within the episode. The \texttt{===} dividers are not part of the model input.

\paragraph{Crop-growth task (WOFOST).} Potato season, profit objective, think variant, full N/P/K/water menu.

\begin{Verbatim}[fontsize=\footnotesize,breaklines=true,breakanywhere=true,breaksymbolleft={}]
=== SYSTEM ===
You are an agricultural management expert. Optimize the management objective specified in the user message. Use only the available actions and respond in the required format.

=== USER (turn 14) ===
We are growing potato from sow to maturity. The planned growing window spans 180 days, and actions are taken every 10 days. Today is day 140 of the season, corresponding to May 20.

<management objective id="profit_max">
Goal: maximize terminal WSO-equivalent profit, not raw yield.

Score at the end of the season:
profit_wso_kg_ha =
  final_WSO_kg_ha
  - 3.5 * total_N_kg_ha
  - 5.5 * total_P_kg_ha
  - 2.25 * total_K_kg_ha
  - 0.5 * total_irrig_cm

Input unit costs:
- Nitrogen: 1 kg/ha N costs 3.5 kg/ha WSO-equivalent.
- Phosphorus: 1 kg/ha P costs 5.5 kg/ha WSO-equivalent.
- Potassium: 1 kg/ha K costs 2.25 kg/ha WSO-equivalent.
- Irrigation: 1 cm water costs 0.5 kg/ha WSO-equivalent.

Successful crop seasons usually produce final yield on the order of thousands of kg/ha.

Decision rule: apply an input only if that input is expected to increase final WSO by more than its WSO-equivalent cost.
</management objective>

Below is the current observation for this step.

<current observation>
[Crop status]
- Finish flag (1 means the season has ended): 0
- Development stage index: 1.136 (DVS=1 indicates flowering; DVS=2 indicates maturity)
- Storage organ dry matter (kg/ha; harvestable yield biomass proxy): 27.17
[Soil nutrients]
- Available soil nitrogen (kg/ha): 8.307
- Available soil phosphorus (kg/ha): 30.83
- Available soil potassium (kg/ha): 0.011
[Soil & water]
- Root-zone soil moisture (fraction): 0.3302
[Cumulative actions]
- Cumulative nitrogen applied so far (kg/ha): 40
- Cumulative phosphorus applied so far (kg/ha): 40
- Cumulative potassium applied so far (kg/ha): 0
- Cumulative irrigation depth applied so far (cm): 0
[Weather]
- Daily solar radiation (J/m²/day): 1.629e+07
- Mean air temperature (°C): 17.03
- Daily rainfall (cm): 0.216
</current observation>

Available actions (pick exactly one):
- Do nothing.
- Apply nitrogen fertilizer (20.0, 40.0, 60.0, or 80.0 kg/ha).
- Apply phosphorus fertilizer (20.0, 40.0, 60.0, or 80.0 kg/ha).
- Apply potassium fertilizer (20.0, 40.0, 60.0, or 80.0 kg/ha).
- Irrigate with 5.0, 10.0, 15.0, or 20.0 cm of water.

Please reason briefly before action about:
1. State grounding: describe the current agronomic state and the main limiting factor based on the current observations
2. Decision: choose the action that best balances final-yield gains against input costs at this step

Keep the reasoning concise and decision-focused. Do not restate the full input.

Respond using the exact format: <tool_call> ... </tool_call> <answer> ... </answer> with no extra text.

Example: <tool_call>[reasoning content]</tool_call> <answer>Apply 20.0 kg/ha nitrogen fertilizer.</answer>

=== ASSISTANT (turn 14) ===
<tool_call>[State grounding: The crop is at DVS 1.136, indicating it is in the flowering stage, with a high storage organ dry matter (27.17 kg/ha), suggesting strong yield potential. Soil nitrogen (8.307 kg/ha) and phosphorus (30.83 kg/ha) are already adequate, and potassium is critically low (0.011 kg/ha), which is essential for tuber development and dry matter accumulation. Root-zone moisture (33.02%) is below optimal, indicating a water stress concern. However, the primary limiting factor is potassium, as its deficiency directly impacts tuber quality and dry matter retention. Nitrogen and phosphorus have been heavily applied (40 kg/ha each), raising input costs without clear yield gains. Irrigation is not currently justified given the moderate moisture and high solar radiation. Decision: Apply 40.0 kg/ha potassium fertilizer, as it is the most likely to boost dry matter and yield without excessive cost, aligning with the objective of maximizing WSO-equivalent profit.]</tool_call> <answer>Apply 40.0 kg/ha potassium fertilizer.</answer>
\end{Verbatim}

\paragraph{Crop-planning task (Cycles).} Maize--soybean rotation, gross-revenue objective, with the crop-traits block enabled. Note the task-specific system prompt, the multi-year trajectory context, and the two-integer (crop, planting-week) action format.

\begin{Verbatim}[fontsize=\footnotesize,breaklines=true,breakanywhere=true,breaksymbolleft={}]
=== SYSTEM ===
You are an agricultural planning expert. Your goal is to maximize long-run gross revenue over a multi-year crop rotation by choosing the crop and planting timing for each year. Use the current soil nitrogen state to make a decision that balances immediate returns with future rotation effects. Gross revenue is yield multiplied by the crop price; there are no production costs in this task. Always emit a valid final action. Keep the reasoning to one short sentence and always close the <tool_call> block before the <answer> block.
Use crop traits as prior agronomic knowledge. Combine them with the current soil nitrogen condition and the long-run rotation objective when deciding which crop to plant next.

=== USER (turn 5) ===
We are making one crop-planning decision for the current year in a multi-year rotation. At each decision, choose the crop to plant and its planting timing. The objective is to maximize cumulative gross revenue across the full rotation horizon, not just this year. You must also choose a planting-week index from 0 to 13, representing the allowable planting window in spring.

Current rotation context:
- current_year: 1993
- years_remaining: 15
Current crop prices in dollars per tonne:
- CornRM.100: 262.91
- SoybeanMG.3: 300.43
Past trajectory summary: 4 years completed; crop counts: CornRM.100=2, SoybeanMG.3=2; cumulative gross revenue: 4645.07.
Recent trajectory (last 3 years):
- year: 1990, crop: SoybeanMG.3, planting_week: 4, yield_tonnes: 3.397, gross_revenue: 1020.59
- year: 1991, crop: CornRM.100, planting_week: 5, yield_tonnes: 6.054, gross_revenue: 1591.78
- year: 1992, crop: SoybeanMG.3, planting_week: 4, yield_tonnes: 3.256, gross_revenue: 978.30

<crop traits>
CornRM.100
- Crop type: cereal grain crop (non-legume, C4).
- Nitrogen role: does not fix nitrogen and generally benefits from a strong soil nitrogen supply.
- Rooting and water: deep-rooted (max root depth about 2.0 m) with relatively high water demand and uptake capacity.
- Temperature and maturity: warm-adapted (optimum development temperature about 28 C) and relatively long-season (maturity thermal time about 2050).
- Rotation role: often a high-demand, high-reward crop that can benefit from following a legume or a higher-nitrogen soil state.

SoybeanMG.3
- Crop type: legume grain crop (C3).
- Nitrogen role: fixes nitrogen, so it is usually less dependent on external soil nitrogen than corn or wheat and can improve rotation value for following cereals.
- Rooting and water: shallower-rooted than corn (max root depth about 0.9 m) with moderate water demand.
- Temperature and maturity: warm-adapted (optimum development temperature about 28 C) and shorter-season than CornRM.100 (maturity thermal time about 1350).
- Rotation role: often a useful nitrogen-building break crop that can support future cereal performance rather than only current-year returns.
</crop traits>

<current observation>
Current soil-state observation:
[Soil nitrogen pool]
- Organic soil nitrogen: 8829
- Profile soil nitrate: 78.85
- Profile soil ammonium: 12.78
[Nitrogen transformations]
- Mineralization: 0.09357
- Immobilization: -0.007777
- Net mineralization: 0.08579
- Ammonium nitrification: 0.1287
[Nitrogen losses]
- N2O from nitrification: 0.000323
- Ammonia volatilization: 0.02389
- Nitrate denitrification: 0.003876
- N2O from denitrification: 0.000331
</current observation>

Available actions (pick exactly one):
Crop index options:
0: CornRM.100
1: SoybeanMG.3
DOY index options: 0 to 13
Decision guidance:
1. Identify the current crop-price regime and soil nitrogen condition.
2. Consider how the crop choice affects both this year's gross revenue and future years in the rotation.
3. Avoid blindly choosing the highest-price crop if it creates poor long-run rotation consequences.
4. Choose the crop and planting or management settings most likely to maximize cumulative gross revenue.
Reason in one short sentence before answering about:
1. The current price regime, soil nitrogen condition, and main agronomic constraint.
2. The likely short-term gross revenue and future-rotation consequences of candidate crop choices.
3. Why the selected action is the best long-run gross-revenue decision.
Keep the reasoning concise and decision-focused. Use at most 35 words. Do not restate the full input. If uncertain, still produce a valid <answer> block.

Respond using the exact format: <tool_call> ... </tool_call> <answer>crop_index, doy_index</answer> with no extra text.
Example: <tool_call>Soybeans likely improve rotation value under the current nitrogen state.</tool_call> <answer>1, 4</answer>
Output integers only inside the answer tag.

=== ASSISTANT (turn 5) ===
<tool_call>Current soil nitrogen is high, favoring corn's yield potential; soybeans would fix nitrogen but reduce short-term revenue. Corn maximizes immediate gross revenue and leverages abundant nitrogen for long-run rotation efficiency. </tool_call> <answer>0, 5</answer>
\end{Verbatim}

\FloatBarrier
\subsection{Per-setting prompt deltas}
\label{app:prompts-examples}

Each benchmark setting is a delta on the same template: one (or, in Tier 3, several) of the three schema blocks change while the rest of the contract stays fixed (Table~\ref{tab:prompt-deltas}).

\begin{table}[H]
  \centering
  \small
  \renewcommand{\arraystretch}{1.15}%
  \begin{tabular}{%
    >{\raggedright\arraybackslash\hyphenpenalty=10000\exhyphenpenalty=10000}p{0.24\textwidth}%
    >{\raggedright\arraybackslash\hyphenpenalty=10000\exhyphenpenalty=10000}p{0.20\textwidth}%
    >{\raggedright\arraybackslash}p{0.44\textwidth}}
    \toprule
    Setting & Changed block & Delta relative to the template above \\
    \midrule
    T1.1 weather-regime & None & Same WOFOST crop-growth schema; only latent weather regime changes. \\
    T1.2 cross-crop traits & Context and observation & Crop name and trait descriptor change; profit objective and action menu stay fixed. \\
    T1.3 price-regime & Task and objective context & Cycles planning prompt replaces the crop-growth prompt; gross-revenue objective uses ID/high-corn/high-soybean prices. \\
    T2.1 observation schema & Observation & Observation fields are dropped, renamed, anonymized, or extended; objective and actions stay fixed. \\
    T2.2 action menu & Action & Menu switches among \texttt{lnpkw}, \texttt{lnpk}, \texttt{lnw}, \texttt{ln}, and \texttt{lw}; unavailable action families are stated explicitly. \\
    T2.3 reward formulation & Objective & Objective block switches among yield, profit, water-stewardship, and nutrient-stewardship definitions. \\
    T3.1 cross-simulator & Observation and action & DSSAT and Cycles use native state fields and action units while preserving objective-first ordering. \\
    T3.2 unified policy & Full schema tuple & Training and test mix tuples of simulator, action menu, and objective. \\
    \bottomrule
  \end{tabular}
  \caption{Per-setting prompt deltas, the compact substitute for printing every full prompt.}
  \label{tab:prompt-deltas}
\end{table}

\FloatBarrier
\section{Reproducibility: training and evaluation}
\label{app:training-hparams}
\label{app:exp-protocol}

\textbf{LLM-RL.} Every LLM-RL run of the main experiments trains \texttt{Qwen/Qwen3-4B-Instruct-2507}~\citep{yang2025qwen3} with GRPO~\citep{shao2024deepseekmath} on VeRL~\citep{sheng2024hybridflow}. Think and no-think variants share every setting and differ only in the response-format contract (Section~\ref{sec:framework-training}), which the dataset \texttt{env\_config} writes into each prompt, so the same training script applies to both. The exploratory backbone and optimizer comparisons of Appendix~\ref{app:t3-backbones} are the only exceptions.

\textbf{NN-PPO.} The fixed-interface baseline trains Stable-Baselines3 PPO on the simulator's native numeric observation with one gradient pass per batch (\texttt{n1}), the update schedule of the LLM arms; an eight-pass variant (\texttt{n8}) makes eight gradient passes over each collected PPO batch and appears only in Appendix~\ref{app:t1-neural}. On Tier~1 both arms have a planned budget of 12{,}800 training episodes per 1{,}600-scenario run: LLM-RL trains two epochs with four rollouts per scenario, and every neural baseline makes eight passes over the scenarios (sixteen for the target-trained action-menu PPO, which has 800 scenarios per menu). LLM-RL runs that the rollout filter stops early (Table~\ref{tab:stop-steps}) complete fewer episodes. \texttt{n1} matches the number of gradient passes over each collected batch; the total number of optimizer steps follows each architecture's batch sizes and is not matched. NN-PPO applies to the Tier~1 settings (Section~\ref{sec:framework-pomdp}); Appendix~\ref{app:bl-nnppo} describes the additional neural baselines. T2.3 runs a fixed 400 steps of 16 prompts over its training rows (1{,}600 for multi-objective training; 1{,}602 materialized rows for yield-only training, its 534-scenario file repeated three times), and T3.2 800 steps of 16 prompts over 3{,}200 rows; all other LLM-RL runs plan 200 steps of 16 scenarios each.

\textbf{Seeds.} Each arm trains three independent seeds; one integer sets both the data-shuffling and the data-loader seed, and the think and no-think arms of a cell share it. The scored seeds are replicates r2--r4 of each cell, with two exceptions: T1.3 was trained afresh as r1--r3, and the Cycles-source think arm of T3.1 uses r2, r3 and r5, because r4 never produced a checkpoint (it aborted on a FlashAttention-3 kernel fault) and r5 was trained with FlashAttention-2. The single runs of the submitted version are not among the scored seeds. Neural baselines train seeds 1--3, except PPO~\texttt{n8} on T1.1 (one seed). The Llama-3.2-3B-Instruct pair~\citep{grattafiori2024llama3} and the optimizer and backbone comparisons of Appendix~\ref{app:t3-backbones} (Qwen2.5-3B-Instruct~\citep{qwen2024qwen25}, Qwen3-8B~\citep{yang2025qwen3}) have one seed per arm and are exploratory.

\textbf{Evaluation.} Each run is scored at its terminal checkpoint on the frozen test suite (Appendix~\ref{app:test-suite}) under deterministic decoding (temperature 0), one repeat per episode, with no checkpoint selection on validation; a run that the top-$p$ rollout filter stops early is scored at its stop step (Table~\ref{tab:stop-steps}). The exploratory Tier~3 training curves (Figure~\ref{fig:tier3}c--e) also score intermediate checkpoints (Appendix~\ref{app:test-suite}). Scripted policies (no-action and the scripted agronomic reference) are rolled out on the same scenarios. Think and no-think are paired by seed and scenario; independently trained arms are compared by the difference of their seed means over the propagated SD $\sqrt{s_a^2+s_b^2}$, where $s_a$ and $s_b$ are the two arms' seed SDs. Three seeds give the direction and spread of an effect and are too few for formal significance tests.

\textbf{GPU dependence of decoding.} Greedy decoding is not bit-reproducible across GPUs: where two tokens are nearly tied, a different GPU model or driver can flip the choice, and the episode then diverges. Scoring used A100, A40 and H200 GPUs. All WOFOST and DSSAT terminal-checkpoint cells of the headline tables come from one scoring pass; the rebuilt Cycles suites and the intermediate checkpoints of the Tier~3 training curves were scored later on H200 and A100 GPUs, and the Tier~1 per-crop and per-regime breakdowns use a later re-scoring of the same terminal checkpoints (Appendix~\ref{app:t12-percrop}). Re-scoring the same checkpoints on the WOFOST and DSSAT targets in that later pass left 68 of 159 trained Qwen no-think seed means bit-identical; on the reward-scored targets, trained seed means moved by at most 0.16 (no-think) and 0.29 (think), at most 1.7 scenario-level standard errors, and the untrained model's by up to 0.62 (3.1 standard errors). Two H200 machines with different drivers differed on 6 of 128 WOFOST episodes of one checkpoint. The released re-scoring records list the GPU of each re-scored run.

\textbf{Compute.} LLM-RL runs use two H200 GPUs (four for T3.2); Appendix~\ref{app:operational-costs} gives the wall-clock cost of a Tier~1 run against the CPU-only neural baseline. Test-suite scoring of one checkpoint takes about 3 to 30 minutes on one GPU, depending on the number of suites. We count GPU-hours as the runtime that W\&B logged for a run times its number of training GPUs. The 90 runs behind the confirmatory tables (30 arms, three seeds each) used 1{,}054 GPU-hours on H200 GPUs, and the exploratory backbone and optimizer runs of Appendix~\ref{app:t3-backbones}, with the earlier legs of resumed runs, used 1{,}080 GPU-hours. The neural baselines behind the reported tables trained on CPU nodes without GPUs for 532 hours of W\&B-recorded run time (57 runs, every leg of a resumed run included): 103 hours on T1.1, 112 on T1.2, 265 on T1.3 and 52 on T2.2. Over the whole project, including preliminary, failed, and replaced runs, the runs logged to W\&B used 7{,}721 GPU-hours, 7{,}109 of them on H200 GPUs (88 on the NVL variant) and 605 on A100 GPUs. These totals leave out test-suite scoring and any work not logged to W\&B.

\begin{table}[t]
  \centering
  \small
  \caption{Essential hyperparameters; settings not listed follow the released VeRL and Stable-Baselines3 configs. The reward-variance rollout filter follows \citet{wang2026ragen2} and is disabled for the T2.3 and T3.2 runs.}
  \label{tab:hparams}
  \begin{minipage}[t]{0.49\textwidth}
    \centering
    \setlength{\tabcolsep}{4pt}
    \renewcommand{\arraystretch}{1.15}
    \begin{tabular}{@{}ll@{}}
      \multicolumn{2}{@{}l}{\textbf{LLM-RL} (GRPO on VeRL)} \\
      \toprule
      backbone            & Qwen3-4B-Instruct-2507 \\
      precision           & bfloat16, FSDP\,1 \\
      optimizer           & AdamW, lr $1{\times}10^{-6}$ \\
      advantage           & GRPO, per-group std norm \\
      PPO clip $\epsilon$ & $0.2$; inner epochs $1$ \\
      mini-batch          & $128$; entropy $0$; KL off \\
      rollout             & vLLM, $16$ prompts $\times$ $4$, temp $1.0$ \\
      max prompt/resp.    & $1024$ / $512$ tokens \\
      reward              & per-turn sum; top-$p$ filter $0.9$ \\
      trainer             & $2$ epochs ($4$ on T2.3, T3.2) \\
      hardware            & $2{\times}$ H200 ($4$ GPUs on T3.2) \\
      \bottomrule
    \end{tabular}
  \end{minipage}
  \hfill
  \begin{minipage}[t]{0.49\textwidth}
    \centering
    \setlength{\tabcolsep}{4pt}
    \renewcommand{\arraystretch}{1.15}
    \begin{tabular}{@{}ll@{}}
      \multicolumn{2}{@{}l}{\textbf{NN-PPO} (Stable-Baselines3)} \\
      \toprule
      policy        & MlpPolicy, $2{\times}64$, tanh \\
      learning rate & $2.5{\times}10^{-4}$ \\
      discount      & $\gamma{=}1.0$; GAE $\lambda{=}0.99$ \\
      PPO clip      & $0.2$ \\
      coefficients  & entropy $0.01$; value $0.5$ \\
      grad clip     & $0.5$ \\
      inner epochs  & $K{=}1$ ($8$ for \texttt{n8}) \\
      envs          & $32$ subproc $\times\,72$ steps \\
      minibatches   & $4$ \\
      dataset epochs & $8$ \\
      \bottomrule
    \end{tabular}
  \end{minipage}
\end{table}

\paragraph{Released assets and licenses.}
\label{app:assets}
The code is released at \url{https://github.com/AgriManager-Framework/AgriManager} under the Apache-2.0 license, with the framework, the training and evaluation scripts, the scripted reference policies, the test-suite generation configurations and seeds, the disjointness audits, and the per-run scoring manifest and per-seed result tables behind every reported table. The Hugging Face organization \url{https://huggingface.co/AgriManager-Framework} hosts the datasets under CC BY 4.0, namely the weather pools, the held-out test suite (\texttt{AgriManager-Framework/agrimanager-test-suite}) with its manifests and audit records, and the per-run result records, together with the trained checkpoints. Checkpoints are released under Apache-2.0, except those fine-tuned from Llama-3.2-3B-Instruct, which keep the Llama 3.2 Community License, and the two fine-tuned from Qwen2.5-3B-Instruct, which keep the Qwen Research License. The simulators and their Gym wrappers are not redistributed: the installer fetches each at a pinned commit, and each keeps its own license (Table~\ref{tab:assets}). The Cycles binary is licensed CC BY-NC-ND 4.0, so its non-commercial terms apply to the Cycles environments. The WOFOST weather data were obtained from the NASA Langley Research Center POWER project, funded through the NASA Earth Science Division.

\begin{table}[t]
  \centering
  \footnotesize
  \setlength{\tabcolsep}{3pt}
  \renewcommand{\arraystretch}{1.12}
  \caption{Existing assets used in this paper, with the version used and the license. Commits are abbreviated; the release documentation (\texttt{THIRD\_PARTY\_NOTICES.md}) gives full hashes. Model licenses are those of the Hugging Face model cards, read on 2026-10-25.}
  \label{tab:assets}
  \begin{tabularx}{\linewidth}{@{}l Y Y Y@{}}
    \toprule
    Asset & Use & Version & License \\
    \midrule
    DSSAT-CSM~\citep{jones2003dssat} & crop simulator & \texttt{dssat-pdi} 4.8.0.24\_2, data v4.8.0.28 & BSD-3-Clause \\
    gym-DSSAT~\citep{gautron2022gymdssat} & Gym wrapper for DSSAT & \texttt{gym\_dssat\_pdi} 63f2c52 (\texttt{gym-dssat} 0.0.8) & BSD-3-Clause \\
    WOFOST / PCSE~\citep{dewit2019wofost} & crop simulator & modified PCSE 5.5.6 inside WOFOSTGym & EUPL 1.1 or later \\
    WOFOSTGym~\citep{solow2025wofostgym} & Gym wrapper for WOFOST & upstream 281f0af plus our patch (fork 253cf54) & MIT \\
    Cycles~\citep{kemanian2024cycles} & crop simulator & v0.12.9-alpha release binary & CC BY-NC-ND 4.0 \\
    CyclesGym~\citep{turchetta2022cyclesgym} & Gym wrapper for Cycles & upstream 44e7dbc plus our patch (fork 2eb8758) & BSD-3-Clause \\
    NASA POWER & WOFOST daily weather & daily data via the PCSE provider & attribution per POWER guide \\
    \midrule
    Qwen3-4B-Instruct-2507~\citep{yang2025qwen3} & backbone of all confirmatory runs & Hugging Face, \texttt{Qwen} org. & Apache-2.0 \\
    Qwen3-8B~\citep{yang2025qwen3} & exploratory backbone & Hugging Face, \texttt{Qwen} org. & Apache-2.0 \\
    Qwen2.5-3B-Instruct~\citep{qwen2024qwen25} & exploratory backbone & Hugging Face, \texttt{Qwen} org. & Qwen Research License \\
    Llama-3.2-3B-Instruct~\citep{grattafiori2024llama3} & exploratory backbone & Hugging Face, \texttt{meta-llama} org. & Llama 3.2 Community License \\
    \midrule
    VeRL~\citep{sheng2024hybridflow} & LLM-RL training & 7522bef (v0.7.0 plus one commit) & Apache-2.0 \\
    vLLM & rollout generation & 0.11.0 & Apache-2.0 \\
    Stable-Baselines3~\citep{raffin2021sb3} & neural baselines & 2.6.0a1 (copy in WOFOSTGym) & MIT \\
    \bottomrule
  \end{tabularx}
\end{table}

\FloatBarrier
\section{Baseline and reference policies}
\label{app:baselines}

This appendix specifies the baseline and reference policies of Section~\ref{sec:benchmark-protocol}: the per-simulator scripted agronomic references that anchor the normalized return, the no-action policy, and the neural baselines.

\FloatBarrier
\subsection{Normalization references and raw returns}
\label{app:exp-normalization}
\label{app:bl-roles}

The normalized return $\tilde{R}$ of Section~\ref{sec:benchmark-protocol} is anchored per target by two scripted policies, both evaluated on the same test scenarios as the policy under test and reused across all policies. \textbf{No-action}, the zero-action control (Appendix~\ref{app:bl-control}), sits at $\tilde{R}{=}0$; it is well defined under every $\Sigma$ in the benchmark. The \textbf{scripted agronomic reference} (per-simulator implementations below) sits at $\pm 1$, since $\tilde{R}$ divides by the absolute gap $\lvert R_{\text{ref}}-R_{\text{no-action}}\rvert$. We use the gap as a denominator where the reference clears no-action by at least a tenth of the largest gain over no-action that a think arm reaches on that target; on every such target the reference lies above no-action. Weather-regime meets this condition on none of its ten crop--split targets: its gap ranges from $-0.148$ to $+0.017$ profit-equivalent t/ha and is negative on eight, so T1.1 cannot be normalized and reports the gain over no-action, $R(\pi)-R_{\text{no-action}}$. Since the gap differs across targets, $\tilde{R}$ compares policies within a target only.

The reference marks codified practice and is uncalibrated against experts or optimal control. On WOFOST it is a conservative schedule: the untrained think model already scores above it on every Tier~2 target (normalized 1.4 to 5.7), so a normalized score of 5 means five times the gap between no-action and that schedule.

Tables~\ref{tab:raw-tier1} and~\ref{tab:raw-tier23} list, for every headline target, the raw return of each trained arm (mean $\pm$ SD over three training seeds), the untrained model under both response formats, both reference returns and the gap; normalized scores follow from Eq.~\eqref{eq:normalized-return}. Invalid-output rates of the trained policies are at most 0.006 on every Tier~1 and Tier~2 seed and target, except two cells of the T2.3 yield-only no-think arm (0.046 and 0.012). The untrained think model is less reliable (up to 0.44 on the anonymized T2.1 variant S5 and 0.15 on T2.3 yield); Appendix~\ref{app:invalid-rates} lists the rates of every setting, and Tier~3 rates appear in Appendix~\ref{app:t3-details}.

\begin{table}[h!]
  \centering
  \scriptsize
  \setlength{\tabcolsep}{3.5pt}
  \caption{Raw test-suite returns, Tier~1. Think and no-think: mean $\pm$ SD over three training seeds. Untrained: the backbone before RL under the think / no-think response format (one run). Reference: the scripted agronomic reference; Gap: reference minus no-action. Units: profit-equivalent t/ha on T1.1 and T1.2; gross revenue on T1.3 (Cycles crop planning; no input-cost term). T1.2 coverage settings share one suite, so their references coincide.}
  \label{tab:raw-tier1}
  \begin{tabular}{@{}lcccccc@{}}
    \toprule
    Target & Think & No-think & Untrained (think / no-think) & No-action & Reference & Gap \\
    \midrule
    \multicolumn{7}{@{}l}{\textit{T1.1 weather-regime (profit-equivalent t/ha; not normalized)}} \\
    chickpea id & $2.060 \pm 0.123$ & $0.025 \pm 0.021$ & $-0.084$ / $-0.430$ & $0.010$ & $-0.002$ & $-0.012$ \\
    chickpea drought & $1.855 \pm 0.362$ & $0.087 \pm 0.127$ & $-0.106$ / $-0.251$ & $0.008$ & $0.000$ & $-0.007$ \\
    chickpea wet & $1.567 \pm 0.079$ & $0.029 \pm 0.011$ & $-0.360$ / $-0.349$ & $0.034$ & $0.026$ & $-0.007$ \\
    chickpea hot & $1.486 \pm 0.250$ & $0.040 \pm 0.047$ & $-0.280$ / $-0.461$ & $0.010$ & $-0.019$ & $-0.029$ \\
    chickpea cold & $0.738 \pm 0.319$ & $0.033 \pm 0.029$ & $-0.513$ / $0.019$ & $0.020$ & $0.021$ & $0.001$ \\
    potato id & $8.551 \pm 0.315$ & $0.096 \pm 0.000$ & $6.183$ / $-0.398$ & $0.096$ & $0.029$ & $-0.067$ \\
    potato drought & $6.832 \pm 0.580$ & $0.091 \pm 0.001$ & $4.540$ / $-0.345$ & $0.090$ & $0.022$ & $-0.068$ \\
    potato wet & $9.623 \pm 0.086$ & $0.111 \pm 0.000$ & $7.931$ / $-0.613$ & $0.111$ & $0.128$ & $0.017$ \\
    potato hot & $10.772 \pm 0.386$ & $0.140 \pm 0.000$ & $9.020$ / $-1.333$ & $0.140$ & $0.064$ & $-0.075$ \\
    potato cold & $7.861 \pm 0.286$ & $0.096 \pm 0.000$ & $6.074$ / $-0.014$ & $0.096$ & $-0.052$ & $-0.148$ \\
    \midrule
    \multicolumn{7}{@{}l}{\textit{T1.2 crop-variety (profit-equivalent t/ha)}} \\
    4ID seen & $2.03 \pm 0.06$ & $0.72 \pm 0.00$ & $0.79$ / $0.43$ & $0.72$ & $1.15$ & $0.43$ \\
    4ID held-out & $1.63 \pm 0.10$ & $0.42 \pm 0.00$ & $0.89$ / $0.39$ & $0.42$ & $0.57$ & $0.14$ \\
    8ID seen & $1.81 \pm 0.05$ & $0.76 \pm 0.08$ & $0.79$ / $0.43$ & $0.72$ & $1.15$ & $0.43$ \\
    8ID held-out & $1.73 \pm 0.18$ & $0.42 \pm 0.01$ & $0.89$ / $0.39$ & $0.42$ & $0.57$ & $0.14$ \\
    16ID seen & $1.96 \pm 0.04$ & $0.71 \pm 0.02$ & $0.79$ / $0.43$ & $0.72$ & $1.15$ & $0.43$ \\
    16ID held-out & $1.83 \pm 0.27$ & $0.40 \pm 0.04$ & $0.89$ / $0.39$ & $0.42$ & $0.57$ & $0.14$ \\
    \midrule
    \multicolumn{7}{@{}l}{\textit{T1.3 price-regime (gross revenue)}} \\
    all regimes & $29{,}826 \pm 257$ & $29{,}834 \pm 1{,}064$ & $13{,}752$ / $20{,}451$ & $10{,}891$ & $28{,}555$ & $17{,}664$ \\
    \bottomrule
  \end{tabular}
\end{table}

\begin{table}[h!]
  \centering
  \scriptsize
  \setlength{\tabcolsep}{3.5pt}
  \caption{Raw test-suite returns, Tiers~2 and~3, laid out as Table~\ref{tab:raw-tier1}. WOFOST Tier~2 and T3.2 objectives are divided by a reference yield (per scenario in T2.1--T2.3, fixed at 8{,}500\,kg/ha in T3.2), so the same objective has different scales across the two settings. DSSAT and Cycles profit are in profit-equivalent t/ha; DSSAT and Cycles yield in kg/ha. Cycles rows use the rebuilt Cycles suites (Appendix~\ref{app:test-suite}).}
  \label{tab:raw-tier23}
  \begin{tabular}{@{}lcccccc@{}}
    \toprule
    Target & Think & No-think & Untrained (think / no-think) & No-action & Reference & Gap \\
    \midrule
    \multicolumn{7}{@{}l}{\textit{T2.1 observation schema}} \\
    S1 & $5.74 \pm 0.19$ & $1.16 \pm 0.07$ & $3.43$ / $-0.22$ & $0.40$ & $1.39$ & $0.98$ \\
    S2a & $5.46 \pm 0.23$ & $0.95 \pm 0.23$ & $3.25$ / $-0.80$ & $0.40$ & $1.39$ & $0.98$ \\
    S2b & $1.33 \pm 1.38$ & $0.62 \pm 0.29$ & $1.78$ / $-0.27$ & $0.40$ & $1.39$ & $0.98$ \\
    S2c & $4.69 \pm 1.15$ & $0.96 \pm 0.38$ & $2.42$ / $0.42$ & $0.40$ & $1.39$ & $0.98$ \\
    S2d & $5.67 \pm 0.09$ & $1.16 \pm 0.07$ & $3.92$ / $-0.29$ & $0.40$ & $1.39$ & $0.98$ \\
    S3 & $5.22 \pm 0.30$ & $0.95 \pm 0.29$ & $3.04$ / $-0.41$ & $0.40$ & $1.39$ & $0.98$ \\
    S4 & $4.99 \pm 0.10$ & $1.11 \pm 0.09$ & $2.63$ / $-1.06$ & $0.40$ & $1.39$ & $0.98$ \\
    S5 & $5.35 \pm 0.08$ & $0.81 \pm 0.26$ & $2.54$ / $0.40$ & $0.40$ & $1.39$ & $0.98$ \\
    \midrule
    \multicolumn{7}{@{}l}{\textit{T2.2 action menu (Family A / B)}} \\
    A \texttt{ln} & $5.71 \pm 0.19$ & $1.00 \pm 1.47$ & $5.17$ / $3.57$ & $0.15$ & $1.04$ & $0.89$ \\
    A \texttt{lnpkw} & $5.67 \pm 0.20$ & $1.15 \pm 0.04$ & $3.46$ / $-0.22$ & $0.40$ & $1.39$ & $0.98$ \\
    A \texttt{lnw} & $5.76 \pm 0.30$ & $1.31 \pm 1.13$ & $5.17$ / $1.46$ & $0.12$ & $2.22$ & $2.10$ \\
    A \texttt{lw} & $7.08 \pm 0.12$ & $4.03 \pm 2.19$ & $5.43$ / $2.24$ & $2.77$ & $4.07$ & $1.30$ \\
    B \texttt{ln} & $5.15 \pm 1.31$ & $1.08 \pm 1.04$ & $5.17$ / $3.57$ & $0.15$ & $1.04$ & $0.89$ \\
    B \texttt{lnpkw} & $4.05 \pm 1.82$ & $0.63 \pm 0.02$ & $3.46$ / $-0.22$ & $0.40$ & $1.39$ & $0.98$ \\
    B \texttt{lnw} & $4.25 \pm 1.66$ & $0.14 \pm 0.00$ & $5.17$ / $1.46$ & $0.12$ & $2.22$ & $2.10$ \\
    B \texttt{lw} & $7.01 \pm 0.25$ & $7.08 \pm 0.28$ & $5.43$ / $2.24$ & $2.77$ & $4.07$ & $1.30$ \\
    \midrule
    \multicolumn{7}{@{}l}{\textit{T2.3 reward formulation (multi-objective / yield-only training)}} \\
    multi nutrient & $0.73 \pm 0.11$ & $-2.40 \pm 0.77$ & $-1.01$ / $-2.40$ & $-2.87$ & $-2.17$ & $0.70$ \\
    multi profit & $0.68 \pm 0.04$ & $0.05 \pm 0.00$ & $0.40$ / $-0.02$ & $0.05$ & $0.16$ & $0.12$ \\
    multi water & $0.67 \pm 0.12$ & $-1.38 \pm 1.47$ & $0.14$ / $-2.15$ & $-2.23$ & $-1.76$ & $0.46$ \\
    multi yield & $0.87 \pm 0.02$ & $0.24 \pm 0.30$ & $0.51$ / $0.04$ & $0.05$ & $0.26$ & $0.21$ \\
    yield-only nutrient & $0.34 \pm 0.63$ & $-0.07 \pm 0.60$ & $-1.01$ / $-2.40$ & $-2.87$ & $-2.17$ & $0.70$ \\
    yield-only profit & $0.60 \pm 0.13$ & $0.51 \pm 0.06$ & $0.40$ / $-0.02$ & $0.05$ & $0.16$ & $0.12$ \\
    yield-only water & $0.41 \pm 0.27$ & $0.67 \pm 0.10$ & $0.14$ / $-2.15$ & $-2.23$ & $-1.76$ & $0.46$ \\
    yield-only yield & $0.91 \pm 0.01$ & $0.87 \pm 0.03$ & $0.51$ / $0.04$ & $0.05$ & $0.26$ & $0.21$ \\
    \midrule
    \multicolumn{7}{@{}l}{\textit{T3.1 cross-simulator, source $\to$ target (profit-equivalent t/ha)}} \\
    Cycles $\to$ Cycles & $11.35 \pm 0.01$ & $11.39 \pm 0.09$ & $10.71$ / $10.31$ & $4.27$ & $10.83$ & $6.55$ \\
    Cycles $\to$ DSSAT & $3.05 \pm 1.35$ & $1.54 \pm 0.07$ & $1.07$ / $1.76$ & $1.07$ & $7.36$ & $6.28$ \\
    Cycles $\to$ WOFOST & $4.21 \pm 0.69$ & $2.89 \pm 0.77$ & $5.03$ / $1.46$ & $0.12$ & $2.22$ & $2.10$ \\
    DSSAT $\to$ Cycles & $9.87 \pm 0.44$ & $8.54 \pm 0.37$ & $10.71$ / $10.31$ & $4.27$ & $10.83$ & $6.55$ \\
    DSSAT $\to$ DSSAT & $6.97 \pm 0.68$ & $7.78 \pm 0.25$ & $1.07$ / $1.76$ & $1.07$ & $7.36$ & $6.28$ \\
    DSSAT $\to$ WOFOST & $4.60 \pm 0.21$ & $0.68 \pm 0.45$ & $5.03$ / $1.46$ & $0.12$ & $2.22$ & $2.10$ \\
    WOFOST $\to$ Cycles & $10.11 \pm 0.43$ & $9.95 \pm 1.92$ & $10.71$ / $10.31$ & $4.27$ & $10.83$ & $6.55$ \\
    WOFOST $\to$ DSSAT & $3.91 \pm 1.65$ & $1.38 \pm 0.33$ & $1.07$ / $1.76$ & $1.07$ & $7.36$ & $6.28$ \\
    WOFOST $\to$ WOFOST & $6.41 \pm 0.07$ & $6.05 \pm 0.02$ & $5.03$ / $1.46$ & $0.12$ & $2.22$ & $2.10$ \\
    \midrule
    \multicolumn{7}{@{}l}{\textit{T3.2 unified policy (held-out: WOFOST \texttt{lnw} profit, \texttt{lnpkw} nutrient, DSSAT and Cycles profit)}} \\
    Cycles profit & $10.55 \pm 0.30$ & $5.89 \pm 3.56$ & $10.68$ / $10.31$ & $4.27$ & $10.83$ & $6.55$ \\
    Cycles yield & $12{,}235 \pm 0$ & $12{,}235 \pm 0$ & $12{,}235$ / $12{,}235$ & $4{,}272$ & $11{,}403$ & $7{,}131$ \\
    DSSAT profit & $6.16 \pm 0.85$ & $4.87 \pm 3.37$ & $1.07$ / $1.74$ & $1.07$ & $7.36$ & $6.28$ \\
    DSSAT yield & $8{,}783 \pm 1{,}646$ & $7{,}564 \pm 844$ & $1{,}796$ / $5{,}925$ & $1{,}074$ & $7{,}778$ & $6{,}704$ \\
    WOFOST \texttt{lnpk} profit & $0.63 \pm 0.04$ & $0.52 \pm 0.39$ & $0.55$ / $-0.06$ & $0.07$ & $0.14$ & $0.07$ \\
    WOFOST \texttt{lnpkw} nutrient & $-0.10 \pm 0.95$ & $-1.20 \pm 2.45$ & $-1.02$ / $-2.41$ & $-2.87$ & $-2.17$ & $0.70$ \\
    WOFOST \texttt{lnpkw} profit & $0.59 \pm 0.12$ & $0.24 \pm 0.22$ & $0.39$ / $-0.03$ & $0.05$ & $0.16$ & $0.12$ \\
    WOFOST \texttt{lnw} profit & $0.53 \pm 0.25$ & $0.18 \pm 0.12$ & $0.60$ / $0.17$ & $0.01$ & $0.26$ & $0.25$ \\
    WOFOST \texttt{lnw} yield & $0.78 \pm 0.07$ & $0.69 \pm 0.16$ & $0.57$ / $0.31$ & $0.01$ & $0.44$ & $0.43$ \\
    \bottomrule
  \end{tabular}
\end{table}

\FloatBarrier
\subsection{WOFOST scripted agronomic reference}
\label{app:bl-wofost}

The WOFOST scripted agronomic reference (used on T1.1, T1.2, T2.1, T2.2, T2.3, the WOFOST target of T3.1 and the WOFOST tuples of T3.2) is a crop-specific rule set. It assigns each crop a set of seasonal input caps and uses crop stage and soil observations to select an action. It is the version behind the submitted figures, re-run on the test suite.

\paragraph{Agronomic sources.}
We designed the nutrient-management rules using the 4R framework and soil-test-based extension recommendations~\citep{bruulsema20124r,franzen2010sf882}. The crop-specific rules draw on recommendations for inoculated pulse legumes~\citep{saskatchewan_chickpea,franzen2023pulses}, potato nutrient management~\citep{franzen2021potato}, and irrigated maize~\citep{davis1996corn}. The irrigation rules use soil moisture and crop stage following FAO-56 and CROPWAT~\citep{allen1998fao56,smith1992cropwat}, with a tighter potato trigger during tuber initiation and bulking~\citep{usu_potato_irrigation}. These references guide the rule structure and qualitative constraints; Table~\ref{tab:ag-heuristic-records} and the decision rule below specify the exact policy configuration used in our experiments.

\paragraph{Per-crop records.}
Each crop resolves to a record of seasonal caps; crops absent from the table use the default row. All seven pulse legumes have both their seasonal N cap and starter-N allowance set to zero.

\begin{table}[t]
\centering
\small
\caption{Seasonal caps of the WOFOST scripted agronomic reference. N/P/K are in kg\,ha$^{-1}$ and irrigation is in mm. Rows not listed use the default. The seven pulse legumes (chickpea, cowpea, faba bean, groundnut, mung bean, pigeon pea, and soybean) use $N=0$ with a zero starter allowance.}
\label{tab:ag-heuristic-records}
\begin{tabular}{lrrrrr}
\toprule
Crop & $N$ & $P$ & $K$ & Irrig. & Trigger \\
\midrule
default      & 60  & 20 & 20 & 80  & 0.31 \\
barley       & 90  & 20 & 20 & 80  & 0.31 \\
cotton       & 80  & 20 & 40 & 120 & 0.31 \\
maize        & 120 & 20 & 20 & 100 & 0.31 \\
millet       & 60  & 20 & 20 & 60  & 0.31 \\
potato       & 120 & 40 & 80 & 180 & 0.37 \\
rapeseed     & 120 & 20 & 40 & 80  & 0.31 \\
rice         & 100 & 20 & 20 & 160 & 0.31 \\
sorghum      & 70  & 20 & 20 & 70  & 0.31 \\
sugarbeet    & 100 & 20 & 60 & 120 & 0.31 \\
sunflower    & 60  & 20 & 20 & 80  & 0.31 \\
sweetpotato  & 40  & 20 & 40 & 80  & 0.31 \\
wheat        & 100 & 20 & 20 & 80  & 0.31 \\
pulse legume & 0   & 20 & 20 & 60  & 0.29 \\
chickpea     & 0   & 20 & 0  & 50  & 0.28 \\
\bottomrule
\end{tabular}
\end{table}

The default P and K soil thresholds are both $5$, and the default minimum temperature for fertilizer application is $4\,^{\circ}$C. Pulse legumes use a minimum temperature of $6\,^{\circ}$C. Potato instead uses P and K thresholds of $7$ and $8$, respectively, and a minimum temperature of $5\,^{\circ}$C.

\paragraph{Objective adjustment.}
The active objective scales the selected crop record. Under \texttt{yield\_max}, all caps are multiplied by $1.15$ and the irrigation trigger is raised by $0.02$. Under \texttt{nutrient\_stewardship}, the N/P/K caps are multiplied by $0.70$. Under \texttt{water\_stewardship}, the irrigation cap is multiplied by $0.60$ and the trigger is lowered by $0.03$. \texttt{profit\_max} uses the record unchanged.

\paragraph{Decision rule.}
Let DVS denote crop development stage, SM soil moisture, and $\mathrm{TOTN}$, $\mathrm{TOTP}$, $\mathrm{TOTK}$, and $\mathrm{TOTIRRIG}$ the cumulative applied amounts. Nitrogen is staged as a fraction of the crop's cap: $0.40N$ for $0.15\le\mathrm{DVS}<0.45$, $0.80N$ for $0.45\le\mathrm{DVS}<0.90$, and the full cap only under \texttt{yield\_max} for $0.90\le\mathrm{DVS}<1.10$. At each decision turn, the policy takes the first applicable rule:

\begin{enumerate}
  \item If rainfall is low, irrigation is exposed, $0.15\le\mathrm{DVS}<1.35$, the seasonal water cap is not spent, and SM is more than $0.04$ below the trigger, irrigate. Acute water stress preempts nutrients because the action menu permits only one action per turn.
  \item Otherwise, if temperature is above the crop's minimum for fertilizer application and $0.05\le\mathrm{DVS}<0.55$, apply P if the available P pool is below the crop's soil-test threshold and $\mathrm{TOTP}$ is below cap; otherwise apply K under the same condition.
  \item Otherwise, if nitrogen is exposed and $0.05\le\mathrm{DVS}<1.10$, apply only the starter allowance to a pulse legume, and only while $\mathrm{DVS}<0.35$. For any other crop, apply nitrogen up to the stage target, unless the available soil-N pool exceeds $25$\,kg\,ha$^{-1}$.
  \item Otherwise, irrigate if rainfall is low and SM is below the trigger.
  \item Otherwise, emit the no-op action.
\end{enumerate}

The selected symbolic action is then projected into the current action menu: if the chosen action family exists, the adapter takes the largest available discrete level that does not exceed the remaining cap; if the family is absent, it emits no-op. This projection lets the same policy operate under both full and restricted WOFOST action schemas.

\FloatBarrier
\subsection{Cycles scripted references}
\label{app:bl-cycles}

\textbf{Crop planning (T1.3).} The action is the annual crop choice from $\{\text{CornRM.100}, \text{SoybeanMG.3}\}$ together with a planting-week index from 0 to 13. The scripted agronomic reference is a \textbf{price-led alternation}: each year it plants the crop with the higher price in the episode's price vector, unless that crop was planted the year before, in which case it plants the other (soybean when the repeat would be corn). With two crops this plants the higher-priced crop in the first year and alternates every year after; it ignores soil state. It plants at the middle week index (7) in the first year and repeats that week every year. The code also defines three diagnostic crop-choice scripts for price responsiveness and rotation sensitivity, which are not scored on the test suite: \emph{price-greedy} plants the higher-priced crop each year; \emph{rotation-aware} plants the higher-priced crop but switches after a repeat when the price dominance ratio is below a crop-specific threshold ($1.35$ for corn, $1.45$ for soybean) and forces a soybean year after two corn years under high-corn prices; \emph{soil-aware} adds a switch to soybean after corn whenever soil-profile NO$_3$ is below $25$\,kg\,ha$^{-1}$.

\textbf{Crop growth (T3.1 and T3.2).} The Cycles maize task exposes nitrogen only, at eleven levels from 0 to 150\,kg/ha. Its reference is the first implementation of our generic scripted policy, which splits nitrogen by season progress (the fraction of the episode's decision turns elapsed): 40\,kg/ha between 15\% and 35\% of the season while cumulative N is below 40\,kg/ha, 60\,kg/ha between 35\% and 65\% while it is below 110\,kg/ha, and, under the yield objective only, 30\,kg/ha between 65\% and 80\% while it is below 160\,kg/ha, each rounded to the nearest available level. This implementation reproduces the submitted reference exactly (10.83 profit-equivalent t/ha on the submitted T3.1 validation scenarios). The current version of the same module has no nitrogen rule on this task; it returns the zero action and scores exactly as no-action, so it is not used.

\FloatBarrier
\subsection{DSSAT scripted schedule}
\label{app:bl-dssat}

The DSSAT reference is a fixed schedule applied on exact days after planting (DAP) inside the daily DSSAT loop, so scheduled events are not skipped by the agent's decision interval. It applies nitrogen at DAP $40$ ($27$\,kg/ha), DAP $45$ ($35$\,kg/ha), and DAP $80$ ($54$\,kg/ha), and irrigation at DAPs $6, 20, 37, 50, 54, 65, 69, 72, 75, 77, 80, 84, 91, 101, 104, 105$, for a total of $264$\,mm. It is the DSSAT scripted agronomic reference for the DSSAT target of T3.1 and the DSSAT tuples of T3.2.

\paragraph{Provenance of the DSSAT reference.} The DSSAT reference values in the submitted version were set by hand. The first implementation matched schedule days exactly and, at the agent's decision interval, applied no input, so it scored as no-action; the submitted figures replaced its values with a manual correction for which no run or code survives. Every DSSAT number in this version uses a reconstruction of the schedule above from code. It reproduces the submitted input totals (116\,kg/ha N, 264\,mm irrigation) and returns 7.17 profit-equivalent t/ha on the submitted validation scenarios, against the hand-set 7.47; on the test suite it returns 7.36.

\FloatBarrier
\subsection{No-action policy}
\label{app:bl-control}

The no-action policy emits $\text{action}_t\!=\!0$ for discrete action spaces and the zero vector for continuous spaces; in WOFOST, DSSAT and Cycles crop growth this applies no fertilizer and no irrigation. On T1.3 crop planning the zero action is crop index~0 at planting-week index~0, so no-action there plants corn in the earliest week every year, which still earns gross revenue. The random-action control of the submitted figures is not scored on the test suite.

\FloatBarrier
\subsection{NN-PPO architecture}
\label{app:bl-nnppo}

The fixed-interface NN baseline is Stable-Baselines3 PPO~\citep{raffin2021sb3} with the stock \texttt{MlpPolicy}: two hidden layers of $64$ units with $\tanh$ activations and separate actor and critic towers that read the same numeric observation, the simulator's native state vector exposed by the env's \texttt{nn\_adapter} (Appendix~\ref{app:adapter-contract}); no learned encoder is added. The action head is a categorical distribution over the discrete action menu, matching the cardinality declared in $\Sigma_A$. Vectorized environments use $32$ parallel workers with $72$-step rollouts; full optimization hyperparameters are in Table~\ref{tab:hparams}. The fixed-interface PPO applies to the Tier~1 settings, whose observation layout and action set match training (Section~\ref{sec:framework-pomdp}); the main text reports one gradient pass per batch (\texttt{n1}), and Table~\ref{tab:t1-neural} adds the eight-pass variant (\texttt{n8}; three seeds on T1.2 and one seed on T1.1, the latter exploratory).

\paragraph{Additional neural baselines.} On potato weather-regime, Stable-Baselines3 DQN and A2C train on the same observation, action menu and 12{,}800-episode budget; A2C keeps the settings it shares with PPO and its own default learning rate ($7{\times}10^{-4}$ with RMSprop). On the action menu, MaskablePPO (sb3-contrib) acts over a canonical 17-action vocabulary (no-op plus four levels each of N, P, K and water) with a legal-action mask from each native menu, trained per family on that family's training menus; a target-trained PPO trains on each test menu alone (800 scenarios, sixteen passes) as a reachable learned reference for that menu. On the test suite, the Family~A think policy scores above masked PPO on every menu (3.1 to 14.4 propagated SD) and above the target-trained PPO on \texttt{lnpkw} (9.7) and \texttt{lnw} (1.6), within one SD of it on \texttt{ln} (1.0) and \texttt{lw} ($-0.08$); the Family~B think policy stays within one propagated SD of both baselines on every menu.

\begin{table}[t]
  \centering
  \footnotesize
  \setlength{\tabcolsep}{4pt}
  \caption{T2.2 action menu: think policies against the schema-capable neural baselines, raw test-suite \texttt{profit\_max} return divided by the per-scenario reference yield (mean $\pm$ SD over three training seeds). Masked PPO trains on its family's two training menus; the target-trained PPO trains on the test menu itself. References for each menu are in Table~\ref{tab:raw-tier23}.}
  \label{tab:t22-neural}
  \begin{tabular}{@{}lccccc@{}}
    \toprule
    & \multicolumn{2}{c}{Family~A (\texttt{lnpkw}, \texttt{lnpk})} & \multicolumn{2}{c}{Family~B (\texttt{lnpk}, \texttt{lw})} & \\
    \cmidrule(lr){2-3}\cmidrule(lr){4-5}
    Test menu & Think & Masked PPO & Think & Masked PPO & Target-trained PPO \\
    \midrule
    \texttt{lnpkw} & $5.67 \pm 0.20$ & $3.18 \pm 0.13$ & $4.05 \pm 1.82$ & $2.24 \pm 0.01$ & $3.16 \pm 0.16$ \\
    \texttt{lnw} & $5.76 \pm 0.30$ & $3.57 \pm 0.21$ & $4.25 \pm 1.66$ & $4.08 \pm 0.14$ & $5.28 \pm 0.10$ \\
    \texttt{ln} & $5.71 \pm 0.19$ & $5.03 \pm 0.11$ & $5.15 \pm 1.31$ & $4.77 \pm 0.07$ & $5.45 \pm 0.18$ \\
    \texttt{lw} & $7.08 \pm 0.12$ & $5.08 \pm 0.08$ & $7.01 \pm 0.25$ & $6.91 \pm 0.08$ & $7.09 \pm 0.08$ \\
    \bottomrule
  \end{tabular}
\end{table}

\FloatBarrier
\section{Additional results}
\label{app:full-results}

This appendix gives the per-seed results behind the main-text statements, the neural baselines on Tier~1, invalid-output rates, the early-stopping record, the training curves, and Tier~3 details. Raw returns and both reference returns for every headline target are in Tables~\ref{tab:raw-tier1} and~\ref{tab:raw-tier23}.

\FloatBarrier
\subsection{Per-seed results}
\label{app:per-seed}

Tables~\ref{tab:seed-tier1} and~\ref{tab:seed-tier23} give every trained seed's test-suite score next to the untrained model. They carry the main-text seed counts: on T2.1 the resource-state variant S2b is the lowest variant in every think seed, and RL leaves think $0.45$ below the untrained model there (mean $0.94$ against $1.39$; one of three seeds above it), while every other variant gains in every seed. On T2.2, Family~A gains over the untrained model on all four menus in every seed; Family~B gains in every seed only on \texttt{lw}. On T2.3, two of the three multi-objective no-think seeds end at or near no-action on every objective.

\begin{table}[h!]
  \centering
  \scriptsize
  \setlength{\tabcolsep}{4pt}
  \caption{Per-seed test-suite scores, Tier~1: normalized return (Eq.~\eqref{eq:normalized-return}) on T1.2 and T1.3, gain over no-action on T1.1. Seeds are listed in training-seed order (r2 / r3 / r4 unless noted). Untrained: one run under the think / no-think response format.}
  \label{tab:seed-tier1}
  \begin{tabular}{@{}lccc@{}}
    \toprule
    Target & Think (per seed) & No-think (per seed) & Untrained (think / no-think) \\
    \midrule
    \multicolumn{4}{@{}l}{\textit{T1.1 weather-regime (gain over no-action, profit-equivalent t/ha)}} \\
    chickpea id & $2.11 / 2.13 / 1.91$ & $0.00 / 0.04 / 0.01$ & $-0.09$ / $-0.44$ \\
    chickpea drought & $2.13 / 1.97 / 1.44$ & $0.00 / 0.23 / 0.01$ & $-0.11$ / $-0.26$ \\
    chickpea wet & $1.62 / 1.47 / 1.51$ & $0.00 / 0.00 / -0.02$ & $-0.39$ / $-0.38$ \\
    chickpea hot & $1.73 / 1.46 / 1.24$ & $0.00 / 0.08 / 0.01$ & $-0.29$ / $-0.47$ \\
    chickpea cold & $1.01 / 0.38 / 0.76$ & $0.00 / 0.05 / -0.01$ & $-0.53$ / $0.00$ \\
    potato id & $8.78 / 8.14 / 8.44$ & $0.00 / 0.00 / 0.00$ & $6.09$ / $-0.49$ \\
    potato drought & $7.41 / 6.39 / 6.43$ & $0.00 / 0.00 / 0.00$ & $4.45$ / $-0.43$ \\
    potato wet & $9.50 / 9.43 / 9.60$ & $0.00 / 0.00 / 0.00$ & $7.82$ / $-0.72$ \\
    potato hot & $10.87 / 10.84 / 10.19$ & $0.00 / 0.00 / 0.00$ & $8.88$ / $-1.47$ \\
    potato cold & $8.07 / 7.51 / 7.71$ & $0.00 / 0.00 / 0.00$ & $5.98$ / $-0.11$ \\
    \midrule
    \multicolumn{4}{@{}l}{\textit{T1.2 crop-variety}} \\
    4ID seen & $3.04 / 3.19 / 2.93$ & $0.00 / 0.00 / 0.00$ & $0.18$ / $-0.66$ \\
    4ID held-out & $7.50 / 8.84 / 8.59$ & $0.00 / 0.00 / 0.00$ & $3.21$ / $-0.21$ \\
    8ID seen & $2.68 / 2.45 / 2.49$ & $0.00 / 0.00 / 0.31$ & $0.18$ / $-0.66$ \\
    8ID held-out & $10.05 / 7.64 / 9.43$ & $0.00 / 0.00 / -0.09$ & $3.21$ / $-0.21$ \\
    16ID seen & $2.94 / 2.96 / 2.78$ & $0.00 / -0.06 / 0.00$ & $0.18$ / $-0.66$ \\
    16ID held-out & $9.61 / 11.63 / 7.85$ & $0.00 / -0.52 / 0.00$ & $3.21$ / $-0.21$ \\
    \midrule
    \multicolumn{4}{@{}l}{\textit{T1.3 price-regime (seeds r1--r3)}} \\
    all regimes & $1.06 / 1.09 / 1.07$ & $1.04 / 1.04 / 1.14$ & $0.16$ / $0.54$ \\
    \bottomrule
  \end{tabular}
\end{table}

\begin{table}[h!]
  \centering
  \scriptsize
  \setlength{\tabcolsep}{4pt}
  \caption{Per-seed normalized test-suite scores, Tiers~2 and~3, laid out as Table~\ref{tab:seed-tier1}. Cycles rows use the rebuilt Cycles suites.}
  \label{tab:seed-tier23}
  \begin{tabular}{@{}lccc@{}}
    \toprule
    Target & Think (per seed) & No-think (per seed) & Untrained (think / no-think) \\
    \midrule
    \multicolumn{4}{@{}l}{\textit{T2.1 observation schema}} \\
    S1 & $5.64 / 5.28 / 5.34$ & $0.69 / 0.76 / 0.83$ & $3.07$ / $-0.64$ \\
    S2a & $5.31 / 5.23 / 4.87$ & $0.60 / 0.30 / 0.76$ & $2.89$ / $-1.22$ \\
    S2b & $0.67 / 2.46 / -0.30$ & $0.00 / 0.56 / 0.09$ & $1.39$ / $-0.68$ \\
    S2c & $3.02 / 4.83 / 5.21$ & $0.12 / 0.78 / 0.80$ & $2.05$ / $0.01$ \\
    S2d & $5.42 / 5.25 / 5.38$ & $0.69 / 0.78 / 0.83$ & $3.57$ / $-0.70$ \\
    S3 & $4.54 / 5.06 / 5.08$ & $0.68 / 0.22 / 0.77$ & $2.68$ / $-0.83$ \\
    S4 & $4.72 / 4.54 / 4.70$ & $0.67 / 0.66 / 0.82$ & $2.26$ / $-1.48$ \\
    S5 & $4.99 / 5.12 / 4.95$ & $0.10 / 0.60 / 0.53$ & $2.17$ / $0.00$ \\
    \midrule
    \multicolumn{4}{@{}l}{\textit{T2.2 action menu (Family A / B)}} \\
    A \texttt{ln} & $6.49 / 6.20 / 6.09$ & $0.00 / 2.87 / 0.00$ & $5.65$ / $3.86$ \\
    A \texttt{lnpkw} & $5.12 / 5.41 / 5.52$ & $0.71 / 0.79 / 0.79$ & $3.10$ / $-0.63$ \\
    A \texttt{lnw} & $2.85 / 2.60 / 2.61$ & $0.26 / 1.19 / 0.24$ & $2.40$ / $0.64$ \\
    A \texttt{lw} & $3.38 / 3.37 / 3.22$ & $0.00 / 2.92 / 0.00$ & $2.05$ / $-0.40$ \\
    B \texttt{ln} & $6.56 / 3.94 / 6.41$ & $0.84 / 2.30 / 0.00$ & $5.65$ / $3.86$ \\
    B \texttt{lnpkw} & $5.51 / 1.81 / 3.79$ & $0.25 / 0.23 / 0.21$ & $3.10$ / $-0.63$ \\
    B \texttt{lnw} & $2.77 / 1.20 / 1.92$ & $0.01 / 0.01 / 0.01$ & $2.40$ / $0.64$ \\
    B \texttt{lw} & $3.23 / 3.10 / 3.48$ & $3.08 / 3.39 / 3.49$ & $2.05$ / $-0.40$ \\
    \midrule
    \multicolumn{4}{@{}l}{\textit{T2.3 reward formulation (multi-objective / yield-only training)}} \\
    multi nutrient & $5.21 / 5.23 / 4.95$ & $0.00 / 0.07 / 1.94$ & $2.66$ / $0.67$ \\
    multi profit & $5.75 / 5.52 / 5.02$ & $0.00 / 0.00 / 0.00$ & $3.03$ / $-0.62$ \\
    multi water & $5.96 / 6.46 / 6.28$ & $0.00 / 0.00 / 5.48$ & $5.09$ / $0.17$ \\
    multi yield & $3.93 / 4.04 / 3.87$ & $0.00 / 0.21 / 2.60$ & $2.21$ / $-0.01$ \\
    yield-only nutrient & $3.54 / 5.19 / 4.99$ & $4.75 / 4.18 / 3.06$ & $2.66$ / $0.67$ \\
    yield-only profit & $5.45 / 5.40 / 3.43$ & $4.57 / 3.84 / 3.48$ & $3.03$ / $-0.62$ \\
    yield-only water & $5.54 / 6.32 / 5.20$ & $6.01 / 6.42 / 6.26$ & $5.09$ / $0.17$ \\
    yield-only yield & $4.13 / 4.15 / 4.06$ & $4.00 / 4.01 / 3.79$ & $2.21$ / $-0.01$ \\
    \midrule
    \multicolumn{4}{@{}l}{\textit{T3.1 cross-simulator, source $\to$ target (Cycles-source think seeds r2, r3, r5)}} \\
    Cycles $\to$ Cycles & $1.08 / 1.08 / 1.08$ & $1.09 / 1.07 / 1.09$ & $0.98$ / $0.92$ \\
    Cycles $\to$ DSSAT & $0.30 / 0.11 / 0.54$ & $0.08 / 0.08 / 0.06$ & $0.00$ / $0.11$ \\
    Cycles $\to$ WOFOST & $2.19 / 2.08 / 1.57$ & $1.66 / 1.36 / 0.93$ & $2.33$ / $0.64$ \\
    DSSAT $\to$ Cycles & $0.78 / 0.92 / 0.86$ & $0.68 / 0.69 / 0.59$ & $0.98$ / $0.92$ \\
    DSSAT $\to$ DSSAT & $0.88 / 0.88 / 1.06$ & $1.10 / 1.02 / 1.08$ & $0.00$ / $0.11$ \\
    DSSAT $\to$ WOFOST & $2.04 / 2.24 / 2.11$ & $0.51 / 0.12 / 0.16$ & $2.33$ / $0.64$ \\
    WOFOST $\to$ Cycles & $0.88 / 0.83 / 0.96$ & $1.04 / 1.03 / 0.53$ & $0.98$ / $0.92$ \\
    WOFOST $\to$ DSSAT & $0.57 / 0.64 / 0.15$ & $0.02 / 0.11 / 0.01$ & $0.00$ / $0.11$ \\
    WOFOST $\to$ WOFOST & $2.95 / 3.02 / 3.01$ & $2.82 / 2.83 / 2.82$ & $2.33$ / $0.64$ \\
    \midrule
    \multicolumn{4}{@{}l}{\textit{T3.2 unified policy}} \\
    Cycles profit & $0.95 / 0.92 / 1.01$ & $0.87 / 0.00 / -0.13$ & $0.98$ / $0.92$ \\
    Cycles yield & $1.12 / 1.12 / 1.12$ & $1.12 / 1.12 / 1.12$ & $1.12$ / $1.12$ \\
    DSSAT profit & $0.77 / 0.70 / 0.96$ & $1.05 / 0.01 / 0.75$ & $0.00$ / $0.11$ \\
    DSSAT yield & $1.28 / 0.87 / 1.31$ & $1.10 / 0.95 / 0.85$ & $0.11$ / $0.72$ \\
    WOFOST \texttt{lnpk} profit & $8.76 / 7.72 / 8.08$ & $8.74 / 0.00 / 10.86$ & $6.91$ / $-1.97$ \\
    WOFOST \texttt{lnpkw} nutrient & $4.60 / 2.40 / 4.86$ & $5.12 / -1.54 / 3.58$ & $2.63$ / $0.66$ \\
    WOFOST \texttt{lnpkw} profit & $5.67 / 3.60 / 4.86$ & $3.77 / 0.00 / 1.19$ & $2.96$ / $-0.63$ \\
    WOFOST \texttt{lnw} profit & $2.72 / 0.93 / 2.65$ & $0.53 / 0.24 / 1.22$ & $2.35$ / $0.63$ \\
    WOFOST \texttt{lnw} yield & $1.84 / 1.59 / 1.92$ & $1.76 / 1.15 / 1.84$ & $1.29$ / $0.70$ \\
    \bottomrule
  \end{tabular}
\end{table}

\FloatBarrier
\subsection{Tier 1 against neural baselines}
\label{app:t1-neural}

Table~\ref{tab:t1-neural} lists the Tier~1 neural baselines. Against PPO with matched updates (\texttt{n1}), the main-text comparison, think leads on every split: by 3.6 to 15.5 propagated SD on weather-regime, 5.6 to 32 on crop-variety and 4.7 on price-regime, with every think seed above every PPO seed. On potato weather-regime it leads A2C, the strongest \texttt{n1} baseline there, by 4.5 to 7.3 propagated SD and DQN by 5.9 to 10.4. PPO with eight passes per batch (\texttt{n8}) makes eight gradient passes over each collected PPO batch. On crop-variety (three seeds) it is ahead of think at 4ID and 8ID ($-0.14$ to $-0.27$ profit-equivalent t/ha, $-0.7$ to $-3.6$ propagated SD) and behind it at 16ID ($+0.08$ and $+0.19$). On weather-regime it was trained with one seed, so its cells are exploratory; that single seed scores above the think mean on nine of the ten crop--split targets, all except potato hot.

\begin{table}[t]
  \centering
  \scriptsize
  \setlength{\tabcolsep}{3.5pt}
  \caption{Tier~1 test-suite returns of think and the neural baselines (mean $\pm$ SD over three training seeds; profit-equivalent t/ha on T1.1 and T1.2, gross revenue on T1.3). PPO~\texttt{n8} on T1.1 is a single seed (exploratory). DQN and A2C train on potato only.}
  \label{tab:t1-neural}
  \begin{tabular}{@{}lccccc@{}}
    \toprule
    Target & Think & PPO \texttt{n1} & PPO \texttt{n8} & A2C & DQN \\
    \midrule
    \multicolumn{6}{@{}l}{\textit{T1.1 weather-regime}} \\
    chickpea id & $2.06 \pm 0.12$ & $0.51 \pm 0.01$ & $2.74$ & -- & -- \\
    chickpea drought & $1.85 \pm 0.36$ & $0.54 \pm 0.07$ & $2.85$ & -- & -- \\
    chickpea wet & $1.57 \pm 0.08$ & $0.17 \pm 0.05$ & $2.16$ & -- & -- \\
    chickpea hot & $1.49 \pm 0.25$ & $0.32 \pm 0.12$ & $2.01$ & -- & -- \\
    chickpea cold & $0.74 \pm 0.32$ & $-0.80 \pm 0.06$ & $1.02$ & -- & -- \\
    potato id & $8.55 \pm 0.32$ & $5.99 \pm 0.17$ & $9.35$ & $6.71 \pm 0.12$ & $-2.15 \pm 1.61$ \\
    potato drought & $6.83 \pm 0.58$ & $3.29 \pm 0.07$ & $7.79$ & $4.08 \pm 0.08$ & $-2.24 \pm 1.42$ \\
    potato wet & $9.62 \pm 0.09$ & $7.42 \pm 0.11$ & $10.32$ & $8.47 \pm 0.13$ & $-1.91 \pm 1.46$ \\
    potato hot & $10.77 \pm 0.39$ & $7.69 \pm 0.33$ & $10.51$ & $7.95 \pm 0.42$ & $-1.53 \pm 1.12$ \\
    potato cold & $7.86 \pm 0.29$ & $5.28 \pm 0.11$ & $8.37$ & $6.46 \pm 0.12$ & $-2.29 \pm 1.60$ \\
    \midrule
    \multicolumn{6}{@{}l}{\textit{T1.2 crop-variety}} \\
    4ID seen & $2.03 \pm 0.06$ & $0.46 \pm 0.05$ & $2.20 \pm 0.00$ & -- & -- \\
    4ID held-out & $1.63 \pm 0.10$ & $0.38 \pm 0.05$ & $1.77 \pm 0.19$ & -- & -- \\
    8ID seen & $1.81 \pm 0.05$ & $0.25 \pm 0.07$ & $2.04 \pm 0.04$ & -- & -- \\
    8ID held-out & $1.73 \pm 0.18$ & $0.41 \pm 0.05$ & $2.01 \pm 0.01$ & -- & -- \\
    16ID seen & $1.96 \pm 0.04$ & $0.13 \pm 0.04$ & $1.88 \pm 0.01$ & -- & -- \\
    16ID held-out & $1.83 \pm 0.27$ & $0.25 \pm 0.06$ & $1.64 \pm 0.21$ & -- & -- \\
    \midrule
    \multicolumn{6}{@{}l}{\textit{T1.3 price-regime}} \\
    all regimes & $29{,}826 \pm 257$ & $28{,}369 \pm 175$ & -- & -- & -- \\
    \bottomrule
  \end{tabular}
\end{table}

\textbf{Lower tail (exploratory).} The later re-scoring of the terminal checkpoints (Appendix~\ref{app:training-hparams}) kept per-episode records for every Tier~1 seed, so we report the mean return of the worst decile of test episodes (CVaR-10), averaged over three seeds. On potato weather-regime, think's CVaR-10 is 2.8 (\texttt{drought}) to 5.4 (\texttt{hot}, \texttt{wet}) profit-equivalent t/ha and 2 of 1{,}920 episodes fall below zero; PPO with matched updates (\texttt{n1}) has $-1.1$ to $2.0$ and 3.7\%. On chickpea, think has $-0.81$ (\texttt{cold}) to $0.66$ and 7.9\% below zero, PPO $-2.4$ to $-1.2$ and 43\%. On crop-variety, think's worst decile is negative in every cell ($-0.56$ to $-0.37$, against about 0 for no-action and $-2.3$ to $-1.8$ for PPO), and 17\% of seen-crop and 33\% of held-out-crop episodes end below no-action on the same scenario. On price-regime no trained policy ends below no-action on any scenario.

\FloatBarrier
\subsection{Tier 1 per-crop breakdown (exploratory)}
\label{app:t12-percrop}

The crop-variety test suite reports the four seen and the four held-out crops as two groups; Table~\ref{tab:t12-percrop} splits it by crop (128 test scenarios per crop, three training seeds per coverage setting). Per-crop splits are not prespecified targets, so the table is exploratory. It gives raw returns of the \texttt{profit\_max} objective in profit-equivalent t/ha (storage-organ yield minus input costs, in grain-equivalent kg/ha, divided by 1{,}000), with no-action and the scripted agronomic reference scored on the same scenarios. Per-crop normalization is not used: the reference gap is unusable on five of the eight crops (below no-action on cotton, millet, groundnut and pigeonpea; 4\% of the largest think gain on seed onion). Per-episode records exist for the later scoring pass of the same terminal checkpoints (Appendix~\ref{app:exp-protocol}), which also supplies the per-regime bars of Figure~\ref{fig:tier1}(c) (pooled over regimes, within $10^{-4}$ of the main-text T1.3 values). Its T1.2 seen and held-out means differ from those of Tables~\ref{tab:raw-tier1} and~\ref{tab:t1-neural} by at most 0.06 t/ha for think, 0.07 for the untrained model and 0.0012 for no-think, and match exactly for PPO and both references.

Think scores above both references on every crop at every coverage except groundnut, where its three means ($-0.16$ to $0.16$ t/ha) lie on both sides of the two references ($-0.03$ and $0.05$). The untrained think model scores 3.74 on seed onion, inside the range of the trained think means (3.42 to 4.21), and below no-action on cotton, millet, groundnut and pigeonpea, where PPO also scores below no-action at every coverage. No-think reproduces no-action exactly at 4ID and stays within 0.20 t/ha of it on every crop at 8ID and 16ID.

\begin{table}[h!]
  \centering
  \scriptsize
  \setlength{\tabcolsep}{2pt}
  \caption{T1.2 crop-variety raw test-suite returns per crop (exploratory), in profit-equivalent t/ha. Think and no-think (seeds r2--r4) and PPO with matched updates (\texttt{n1}, seeds s1--s3): mean $\pm$ SD over three training seeds. Untrained model: one run per response format. No-action and Reference (the scripted agronomic reference): one run, on the same 128 scenarios per crop.}
  \label{tab:t12-percrop}
  \begin{tabular}{@{}lcccccccc@{}}
    \toprule
    & \multicolumn{4}{c}{Seen crops} & \multicolumn{4}{c}{Held-out crops} \\
    \cmidrule(lr){2-5} \cmidrule(lr){6-9}
    Policy & cotton & millet & rapeseed & rice & barley & groundnut & pigeonpea & seed onion \\
    \midrule
    No-action & $0.05$ & $0.06$ & $0.49$ & $2.26$ & $0.58$ & $0.05$ & $0.20$ & $0.87$ \\
    Reference & $-0.05$ & $-0.13$ & $1.24$ & $3.53$ & $1.13$ & $-0.03$ & $0.16$ & $1.01$ \\
    \midrule
    Untrained think & $-0.41$ & $-0.70$ & $1.07$ & $2.95$ & $1.44$ & $-0.77$ & $-0.70$ & $3.74$ \\
    Untrained no-think & $-0.20$ & $0.06$ & $0.49$ & $1.33$ & $0.58$ & $0.00$ & $0.19$ & $0.78$ \\
    \midrule
    Think 4ID & $0.56{\pm}0.16$ & $0.61{\pm}0.19$ & $2.27{\pm}0.25$ & $4.68{\pm}0.10$ & $2.42{\pm}0.06$ & $0.16{\pm}0.15$ & $0.30{\pm}0.17$ & $3.42{\pm}0.40$ \\
    Think 8ID & $0.46{\pm}0.11$ & $0.35{\pm}0.16$ & $2.19{\pm}0.16$ & $4.30{\pm}0.02$ & $2.56{\pm}0.55$ & $-0.16{\pm}0.09$ & $0.33{\pm}0.06$ & $4.21{\pm}0.19$ \\
    Think 16ID & $0.53{\pm}0.05$ & $0.73{\pm}0.16$ & $2.25{\pm}0.18$ & $4.38{\pm}0.14$ & $3.01{\pm}0.71$ & $0.00{\pm}0.06$ & $0.47{\pm}0.04$ & $3.77{\pm}0.58$ \\
    \midrule
    No-think 4ID & $0.05{\pm}0.00$ & $0.06{\pm}0.00$ & $0.49{\pm}0.00$ & $2.26{\pm}0.00$ & $0.58{\pm}0.00$ & $0.05{\pm}0.00$ & $0.20{\pm}0.00$ & $0.87{\pm}0.00$ \\
    No-think 8ID & $0.06{\pm}0.01$ & $0.09{\pm}0.05$ & $0.53{\pm}0.06$ & $2.36{\pm}0.17$ & $0.59{\pm}0.01$ & $0.06{\pm}0.01$ & $0.20{\pm}0.01$ & $0.83{\pm}0.06$ \\
    No-think 16ID & $0.02{\pm}0.06$ & $0.03{\pm}0.05$ & $0.42{\pm}0.12$ & $2.35{\pm}0.16$ & $0.38{\pm}0.34$ & $0.06{\pm}0.02$ & $0.14{\pm}0.11$ & $1.01{\pm}0.25$ \\
    \midrule
    PPO \texttt{n1} 4ID & $-1.29{\pm}0.04$ & $-0.50{\pm}0.06$ & $0.71{\pm}0.11$ & $2.94{\pm}0.11$ & $1.33{\pm}0.09$ & $-0.95{\pm}0.11$ & $-0.96{\pm}0.08$ & $2.09{\pm}0.25$ \\
    PPO \texttt{n1} 8ID & $-1.37{\pm}0.10$ & $-0.73{\pm}0.02$ & $0.56{\pm}0.18$ & $2.53{\pm}0.14$ & $1.28{\pm}0.23$ & $-0.96{\pm}0.06$ & $-0.83{\pm}0.04$ & $2.16{\pm}0.07$ \\
    PPO \texttt{n1} 16ID & $-1.42{\pm}0.06$ & $-0.67{\pm}0.03$ & $0.36{\pm}0.15$ & $2.22{\pm}0.13$ & $0.80{\pm}0.25$ & $-0.92{\pm}0.04$ & $-0.76{\pm}0.10$ & $1.90{\pm}0.06$ \\
    \bottomrule
  \end{tabular}
\end{table}

\FloatBarrier
\subsection{Invalid-output rates}
\label{app:invalid-rates}

Table~\ref{tab:invalid} gives the fraction of decisions whose output the parser cannot map to a legal action, for every setting. A failed decision falls back to the do-nothing action (on T1.3 crop planning, to a scripted crop choice; no T1.3 policy produced an invalid output). The untrained model is scored per target, so its T3.1 entry covers both in-domain and transfer cells.

\begin{table}[t]
  \centering
  \footnotesize
  \setlength{\tabcolsep}{4pt}
  \caption{Invalid-output rate on the test suite, mean (max) over seeds and targets of each setting. Trained arms: three seeds per arm. T3.1 separates in-domain cells (source = target) from the six transfers.}
  \label{tab:invalid}
  \begin{tabular}{@{}lcccc@{}}
    \toprule
    & \multicolumn{2}{c}{Trained} & \multicolumn{2}{c}{Untrained} \\
    \cmidrule(lr){2-3}\cmidrule(lr){4-5}
    Setting & think & no-think & think & no-think \\
    \midrule
    T1.1 weather-regime   & 0.000 (0.000) & 0.000 (0.000) & 0.003 (0.005) & 0.000 (0.000) \\
    T1.2 crop-variety     & 0.001 (0.006) & 0.000 (0.000) & 0.000 (0.000) & 0.000 (0.000) \\
    T1.3 price-regime     & 0.000 (0.000) & 0.000 (0.000) & 0.000 (0.000) & 0.000 (0.000) \\
    T2.1 observation      & 0.000 (0.001) & 0.000 (0.000) & 0.077 (0.438) & 0.000 (0.000) \\
    T2.2 action menu      & 0.000 (0.000) & 0.000 (0.000) & 0.003 (0.006) & 0.000 (0.000) \\
    T2.3 reward           & 0.000 (0.000) & 0.002 (0.046) & 0.048 (0.150) & 0.000 (0.000) \\
    T3.1 in-domain        & 0.000 (0.001) & 0.000 (0.000) & 0.446 (1.000) & 0.000 (0.000) \\
    T3.1 transfers        & 0.113 (0.914) & 0.000 (0.000) & -- & -- \\
    T3.2 unified          & 0.000 (0.003) & 0.000 (0.003) & 0.305 (1.000) & 0.000 (0.000) \\
    \bottomrule
  \end{tabular}
\end{table}

\FloatBarrier
\subsection{Training curves (exploratory)}
\label{app:full-results-training-curves}

The training curves below come from the W\&B logs of the submitted runs, one seed per setting, and are exploratory. They plot the typed environment training signal where available: profit for WOFOST profit-maximization settings, objective reward for schema and reward-shift settings, and normalized gross revenue for Cycles price-regime training. Curve lengths differ across runs: crops and simulators have different season lengths and episode horizons, and the rollout filter stops some LLM-RL runs early (Appendix~\ref{app:no-think-collapse}).

\begin{figure}[H]
  \centering
  \includegraphics[width=0.78\textwidth]{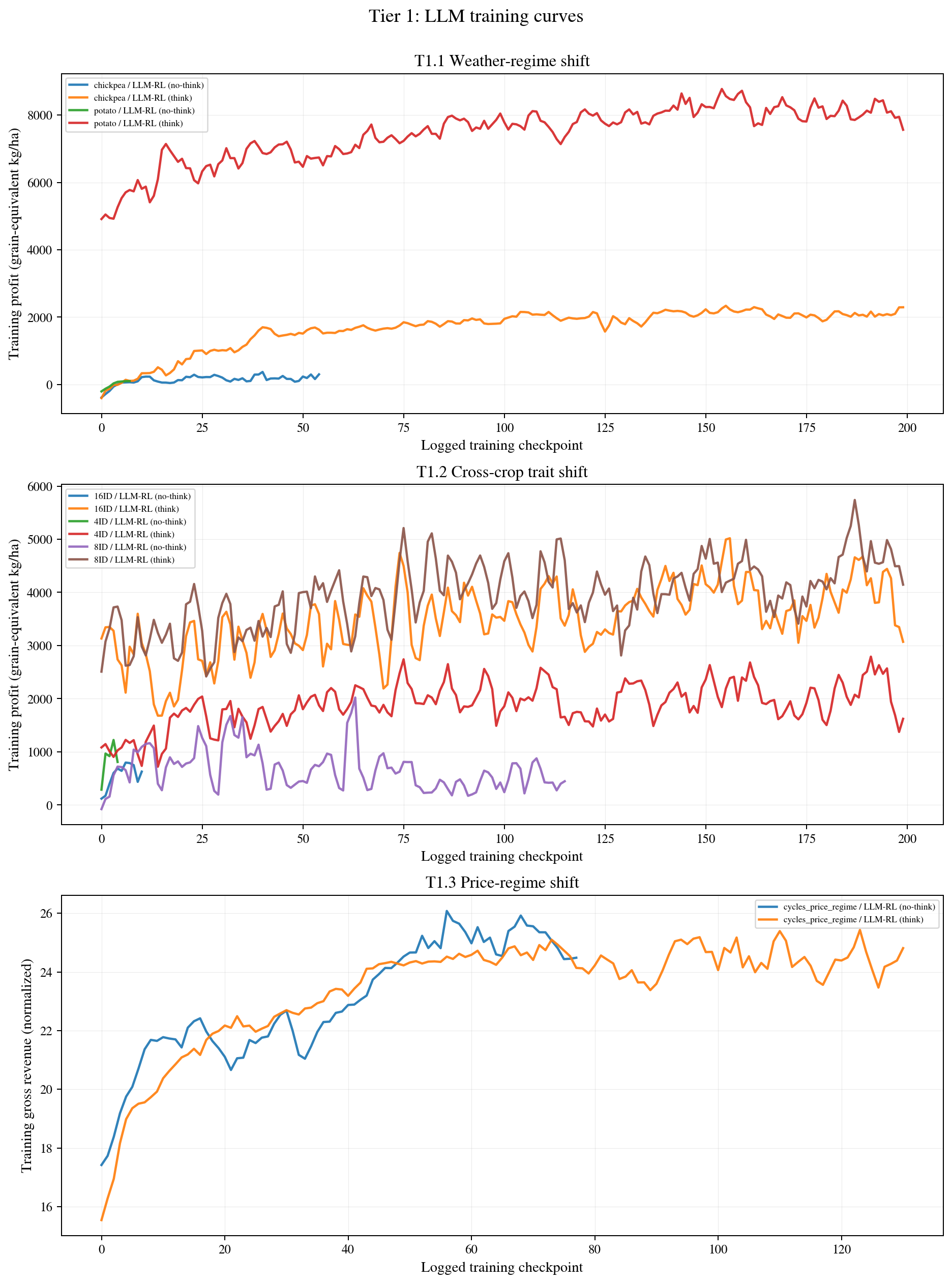}
  \caption{Tier~1 LLM training curves for weather-regime, cross-crop, and price-regime shifts (submitted runs, one seed; exploratory). Each panel uses the typed training metric for that experiment: grain-equivalent profit on WOFOST and normalized gross revenue on Cycles.}
  \label{fig:training-curves-tier1-llm}
\end{figure}

\begin{figure}[H]
  \centering
  \includegraphics[width=\textwidth]{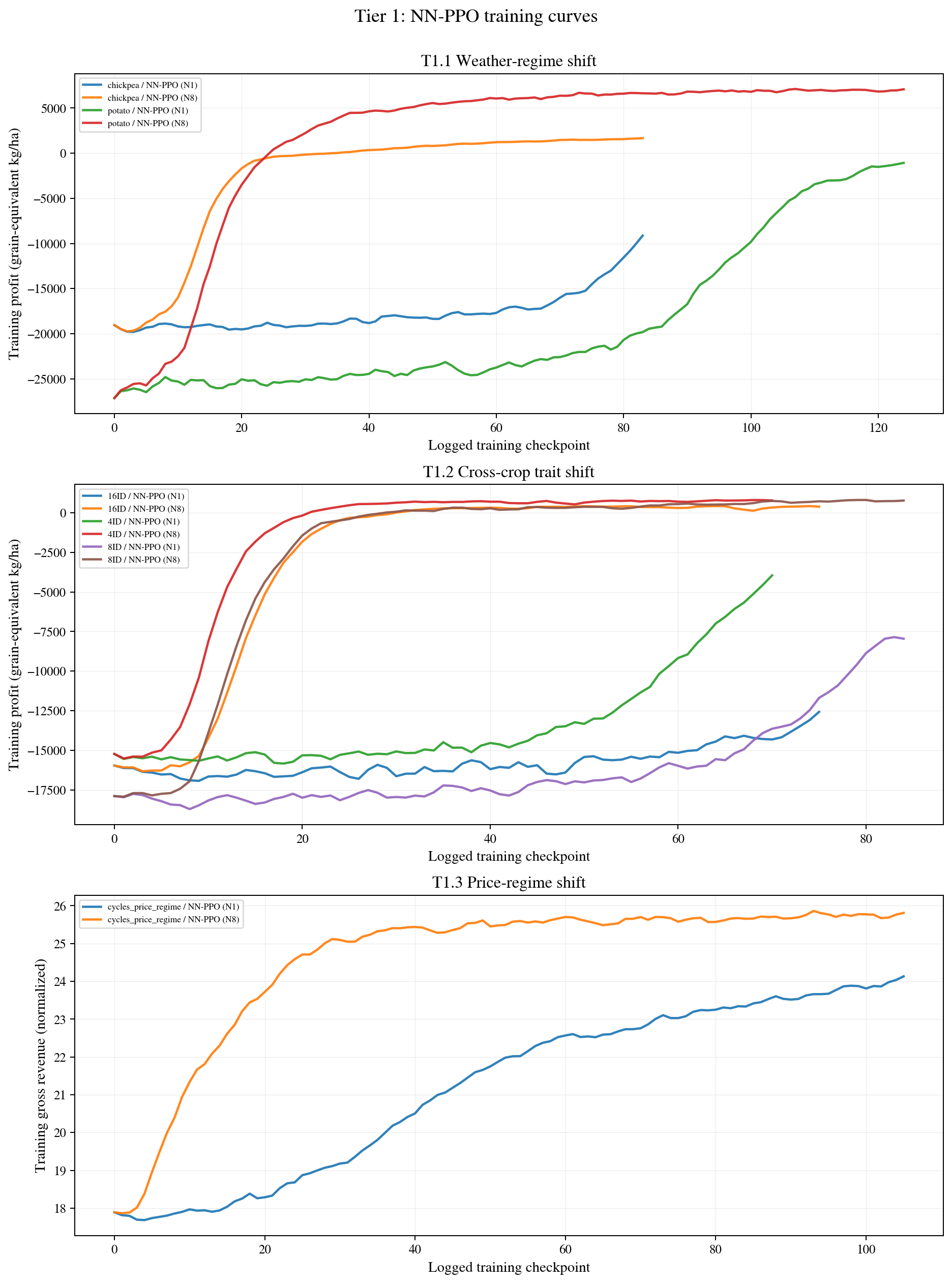}
  \caption{Tier~1 NN-PPO training curves for the same settings (submitted runs, one seed; exploratory).}
  \label{fig:training-curves-tier1-nn}
\end{figure}

\begin{figure}[H]
  \centering
  \includegraphics[width=\textwidth]{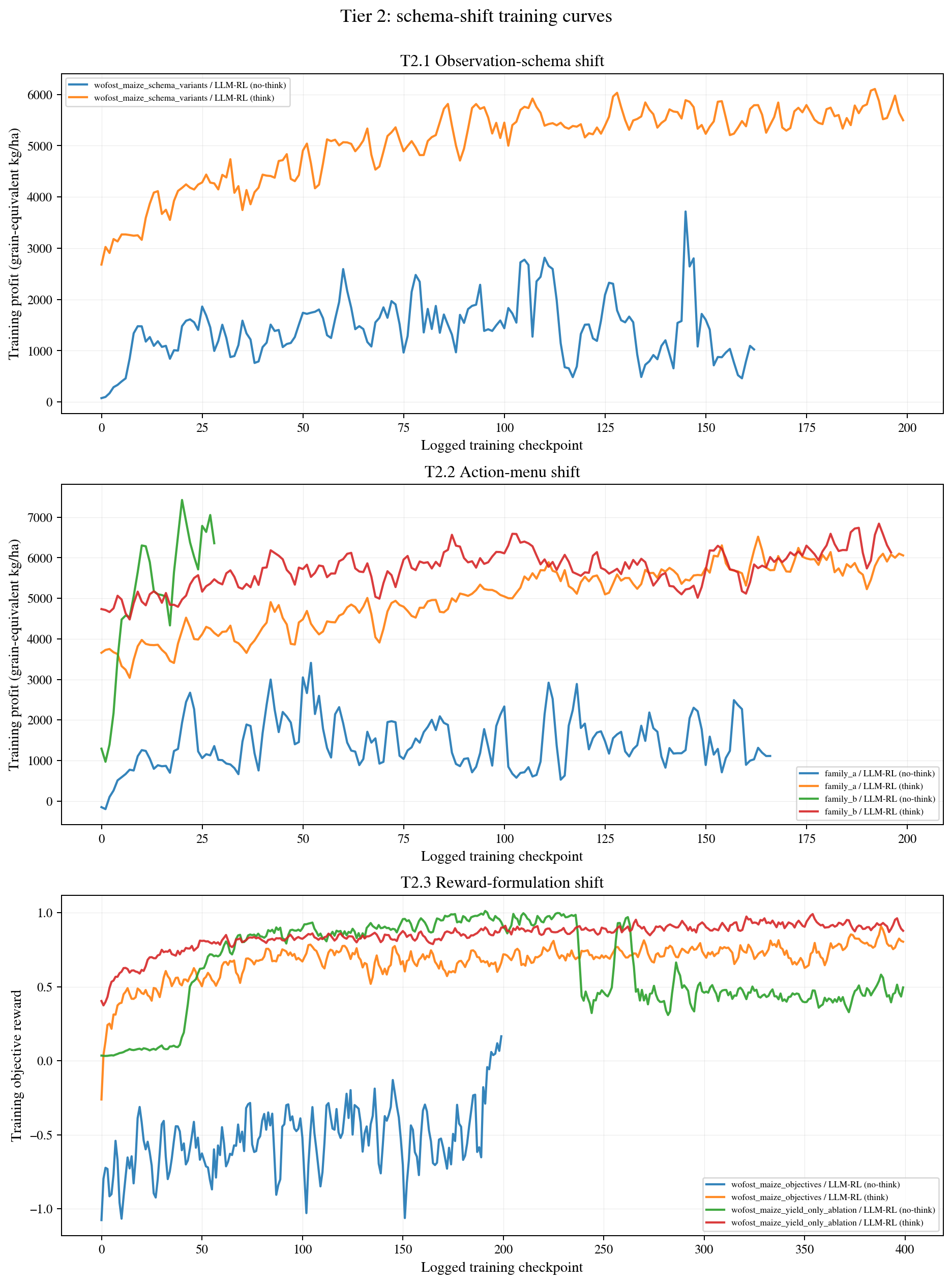}
  \caption{Tier~2 training curves for observation-schema, action-menu, and reward-formulation shifts (submitted runs, one seed; exploratory). The T2.1 panel includes both think and no-think runs; the T2.3 panel includes the multi-objective runs and the yield-only ablation.}
  \label{fig:training-curves-tier2}
\end{figure}

\begin{figure}[H]
  \centering
  \includegraphics[width=\textwidth]{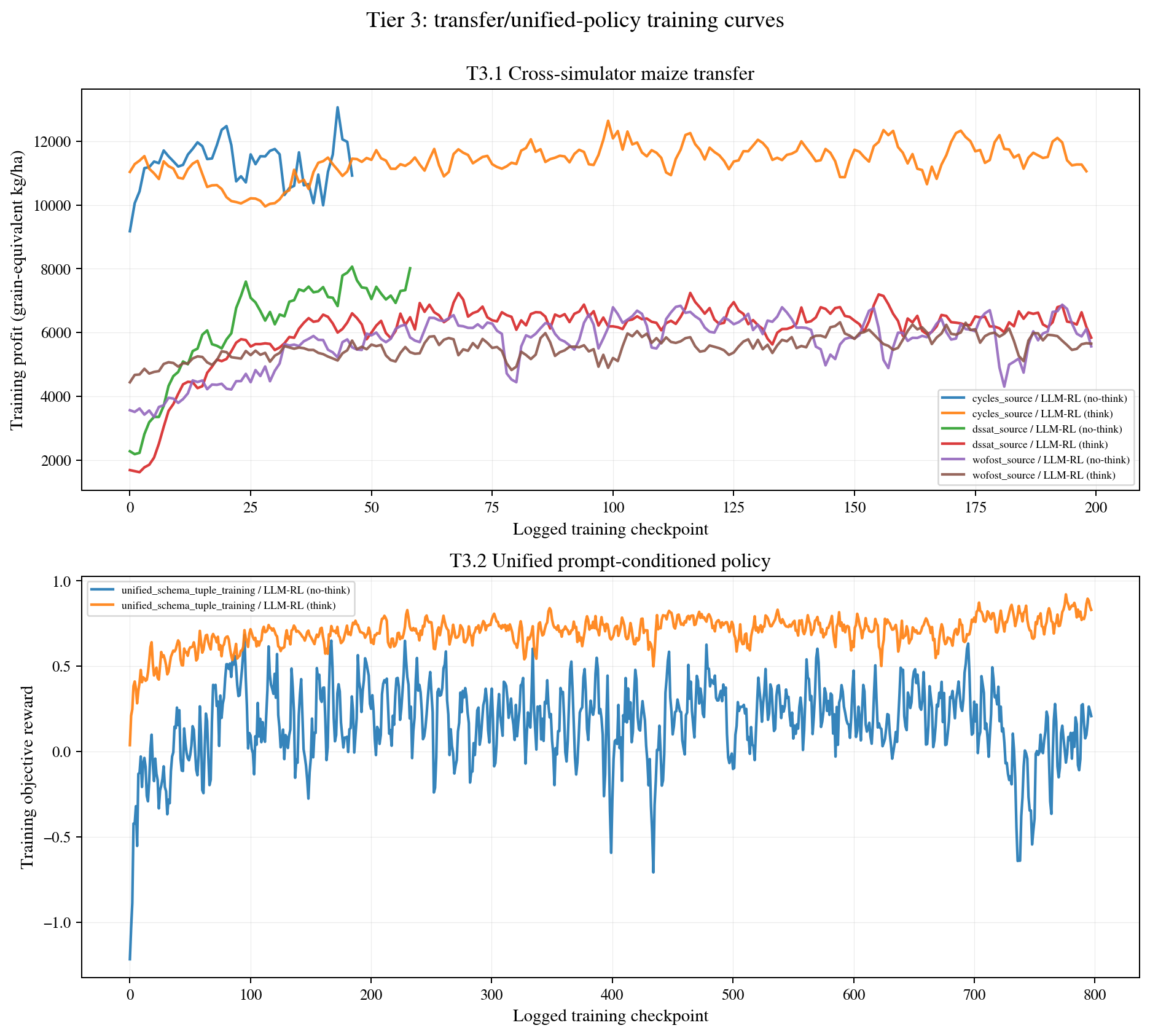}
  \caption{Tier~3 training curves for cross-simulator maize transfer and unified-policy training (submitted runs, one seed; exploratory).}
  \label{fig:training-curves-tier3}
\end{figure}

\FloatBarrier
\subsection{Early stopping under the rollout filter}
\label{app:no-think-collapse}

Training stops early when the top-$p$ rollout filter discards every prompt in a batch for five consecutive steps; a stopped run is scored at its last checkpoint. The filter is on for T1.1, T1.2, T1.3, T2.1, T2.2 and T3.1 and off for T2.3 and T3.2, whose runs all reach their planned 400 and 800 steps. Table~\ref{tab:stop-steps} lists the final step of every scored run. Every think run reaches its planned step except the three T1.3 runs; every no-think run on the two WOFOST Tier~1 settings stops within the first 160 steps, most within 35.

\begin{table}[t]
  \centering
  \footnotesize
  \setlength{\tabcolsep}{4pt}
  \caption{Final training step of each scored run (seeds r2 / r3 / r4 unless noted), out of 200 planned steps. The checkpoint at this step is the one scored on the test suite.}
  \label{tab:stop-steps}
  \begin{tabular}{@{}lcc@{}}
    \toprule
    Cell & Think & No-think \\
    \midrule
    T1.1 chickpea & 200 / 200 / 200 & 18 / 60 / 124 \\
    T1.1 potato & 200 / 200 / 200 & 14 / 17 / 27 \\
    T1.2 4ID & 200 / 200 / 200 & 18 / 33 / 13 \\
    T1.2 8ID & 200 / 200 / 200 & 13 / 13 / 134 \\
    T1.2 16ID & 200 / 200 / 200 & 21 / 160 / 19 \\
    T1.3 (seeds r1 / r2 / r3) & 141 / 154 / 191 & 54 / 43 / 110 \\
    T2.1 & 200 / 200 / 200 & 116 / 88 / 92 \\
    T2.2 Family~A & 200 / 200 / 200 & 99 / 161 / 141 \\
    T2.2 Family~B & 200 / 200 / 200 & 66 / 149 / 187 \\
    T3.1 WOFOST source & 200 / 200 / 200 & 200 / 180 / 200 \\
    T3.1 DSSAT source & 200 / 200 / 200 & 114 / 76 / 78 \\
    T3.1 Cycles source (think r2 / r3 / r5) & 200 / 200 / 200 & 75 / 92 / 65 \\
    \bottomrule
  \end{tabular}
\end{table}

\textbf{Filter mechanism.} We follow the top-$p$ rollout filter of \citet{wang2026ragen2}. For each prompt the trainer samples $G{=}4$ rollouts, ranks prompts by in-group reward standard deviation, and keeps the top $p{=}0.9$ fraction. When all four rollouts of a prompt return identical rewards (zero in-group standard deviation), the prompt carries no signal for GRPO's advantage estimate and is discarded; when every prompt in a batch hits zero variance, the whole batch is filtered.

\textbf{Reward landscape.} WOFOST \texttt{profit\_max} subtracts the unit-cost-weighted sum over four inputs (N, P, K, water) from terminal yield, with $c_N{=}3.5$, $c_P{=}5.5$, $c_K{=}2.25$, $c_W{=}0.5$ (per cm). No-action receives a small but stable positive reward, while any non-zero input is penalized below zero before its yield gain offsets the cost. No-action is therefore a stable basin: once all four rollouts of a prompt converge onto it, the in-group reward variance is zero and the filter removes the prompt.

\begin{figure*}[t]
  \centering
  \includegraphics[width=\linewidth]{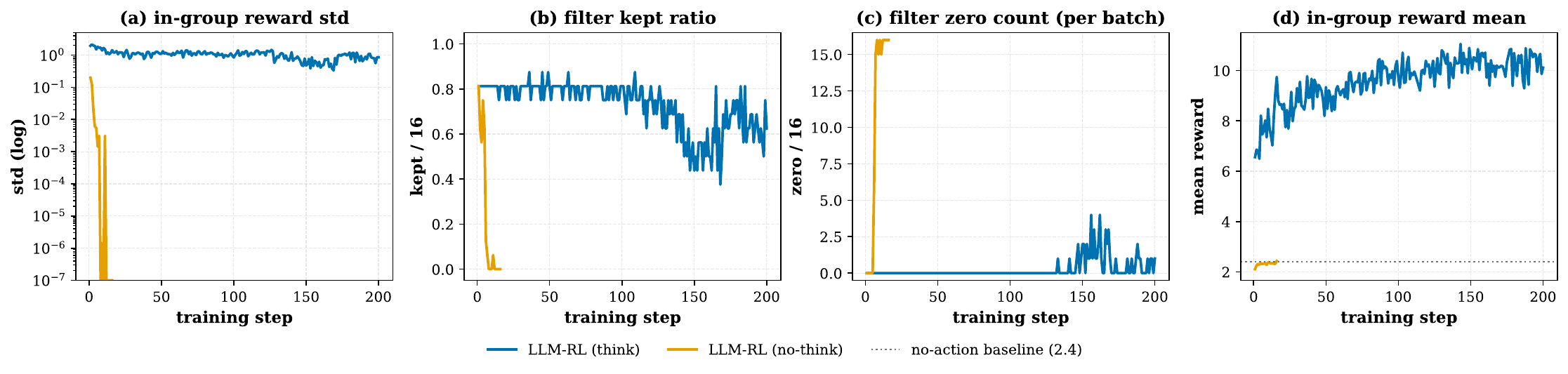}
  \caption{Rollout-level diagnostics across the submitted potato weather-regime training runs (one seed per arm; exploratory). (a)~in-group reward standard deviation (log scale); (b)~fraction of prompts kept by the top-$p$ filter; (c)~number of zero-variance prompts per batch (out of 16); (d)~mean in-group reward, with the no-action baseline marked. The no-think run (orange) stops at step 16; the think run (blue) trains for 200 steps.}
  \label{fig:no-think-collapse}
\end{figure*}

\textbf{Diagnostics from the submitted runs.} Figure~\ref{fig:no-think-collapse} plots four rollout-level diagnostics from the submitted potato weather-regime runs (one seed per arm; exploratory). By step 13 the no-think run's in-group reward standard deviation has fallen to the $10^{-6}$ range, \texttt{filter\_kept\_ratio} is at zero, \texttt{filter\_zero\_count} is at $16/16$, \texttt{consecutive\_all\_filtered\_steps} reaches five, and the run terminates at step 16. The think run holds in-group reward standard deviation above $0.5$ for all 200 steps, keeps \texttt{filter\_kept\_ratio} near $0.6$ on average, and lifts the in-group reward mean to about $10.1$, roughly four times the no-action baseline. The submitted chickpea no-think run shows the same pattern, and the three replicated potato no-think runs stop at steps 14, 17 and 27 (Table~\ref{tab:stop-steps}).

\textbf{A possible reason think keeps training.} This is an observed correlation, and we have not isolated a mechanism. The think variant emits a reasoning trace before the action, so each rollout's token path is longer and its sampling space larger; under the same prompt the four rollouts spread their final actions more widely and the in-group reward standard deviation stays above zero. The no-think variant emits the action directly, so once the policy drifts toward no-action all four rollouts tend to land there together. Under GRPO with a top-$p$ filter the reasoning trace may thus act as a form of exploration. The T1.3 think runs, which also stop early, and the Llama-3.2-3B pair (Appendix~\ref{app:t3-backbones}) show that the effect is not universal.

\FloatBarrier
\subsection{Tier 3 details}
\label{app:t3-details}

\textbf{Unified policy against the target-trained specialist.} Each T3.1 think specialist is also scored on the T3.2 held-out suite of its own simulator, so Table~\ref{tab:t3-specialist} compares the two policies on identical scenarios under the same reward definition, three training seeds each. On the three profit tuples the specialist trains on the evaluated tuple itself, an in-distribution reference, while the unified policy never trains on that tuple and recombines schema components seen in other tuples (zero-shot recombination). On DSSAT profit the unified policy recovers about 86\% of the specialist's gain over the untrained model, with overlapping seeds. On WOFOST \texttt{lnw} profit and Cycles profit every specialist seed scores above every unified seed, and on both the unified mean sits below the untrained model (raw differences $-0.06$ and $-0.13$, with two and one of three seeds above it). On the held-out nutrient tuple, scored against the WOFOST specialist trained on profit, the unified policy is ahead by 0.87 propagated SD.

\begin{table}[t]
  \centering
  \footnotesize
  \setlength{\tabcolsep}{3.5pt}
  \caption{T3.2 unified policy, never trained on these tuples, against the T3.1 specialist trained on the profit tuple itself (on the nutrient row, the WOFOST profit specialist), on the held-out tuples, in normalized return (mean $\pm$ SD over three training seeds; the untrained model is a single run). The last column is the raw-return difference, unified think minus specialist, and that difference in propagated SD. The first three rows are the callback targets of Figure~\ref{fig:tier3}(c--e).}
  \label{tab:t3-specialist}
  \begin{tabular}{@{}lccccc@{}}
    \toprule
    & \multicolumn{2}{c}{Unified} & Specialist & Untrained & Unified $-$ \\
    \cmidrule(lr){2-3}
    Held-out tuple & think & no-think & think & think & specialist \\
    \midrule
    WOFOST \texttt{lnw} profit     & $2.10 \pm 1.01$ & $0.66 \pm 0.50$ & $2.99 \pm 0.06$ & $2.35$ & $-0.22$ ($-0.87$) \\
    DSSAT profit                   & $0.81 \pm 0.13$ & $0.60 \pm 0.54$ & $0.94 \pm 0.11$ & $0.00$ & $-0.81$ ($-0.75$) \\
    Cycles profit                & $0.96 \pm 0.05$ & $0.25 \pm 0.54$ & $1.08 \pm 0.00$ & $0.98$ & $-0.81$ ($-2.70$) \\
    WOFOST \texttt{lnpkw} nutrient & $3.95 \pm 1.35$ & $2.39 \pm 3.49$ & $2.76 \pm 0.13$ & $2.63$ & $+0.83$ ($+0.87$) \\
    \bottomrule
  \end{tabular}
\end{table}

\textbf{Untrained model and format failures.} The untrained think model fails the output format on every DSSAT decision of the profit targets (invalid-output rate 1.00), so it scores as no-action there, and the gains of trained policies over the untrained model on DSSAT profit are gains over no-action. On the DSSAT yield tuple it fails on 93\% of decisions and scores 0.11 in normalized return (1{,}796 against 1{,}074\,kg/ha for no-action). On the Cycles targets it fails on about a third of decisions (0.33 to 0.37) and still reaches 0.98 on profit. Trained transfers also lose part of their score to format failures: across the 18 seeds of the six T3.1 transfers, the invalid-output rate is 0.10 to 0.25 in six seeds and 0.91 in one Cycles$\to$DSSAT seed, against at most 0.001 in the three in-domain cells.

\textbf{Saturated Cycles yield.} On the Cycles yield tuple every Qwen policy (think and no-think unified policies in every seed, and the untrained model under both response formats) and both trained Llama-3.2-3B arms reach the same mean yield, 12{,}235\,kg/ha against 4{,}272 for no-action. This tuple cannot separate competent policies and adds a constant to any seen-control average, one of the cross-simulator confounds of Appendix~\ref{app:t3-inventory}.

\textbf{No-think instability.} Without thinking, the unified Qwen policy ends below think on all four held-out tuples in mean, with think ahead in 10 of 12 seed pairs. Seed r3 of the no-think arm ends near or below no-action on every held-out tuple (normalized 0.01 on DSSAT profit, 0.24 on WOFOST \texttt{lnw} profit, $-1.54$ on nutrient, and exactly no-action on Cycles profit), and seed r4 ends below no-action on Cycles profit ($-0.13$).

\FloatBarrier
\subsection{Backbone dependence of the reasoning result}
\label{app:t3-backbones}

The reading of thinking as a training stabilizer rests on the Qwen backbones. In the author-response period we compared four think and no-think pairs on the T3.2 \emph{validation set}, one training seed per run: Qwen3-4B-Instruct under GRPO, DrGRPO, and PPO, and Qwen2.5-3B-Instruct~\citep{qwen2024qwen25} under GRPO (Table~\ref{tab:t3-retention}). For each run we take the validation reward averaged over the nine seen and held-out tuples at each of its 80 evaluations, find the step where it peaks, and report how much of that peak is lost by the end of training. In all four pairs the think arm peaks late and keeps nearly all of its peak, and the no-think arm peaks early and gives part of it back. These comparisons are exploratory: they use the validation set, one seed per run, and checkpoints chosen by the validation curve itself.

\begin{table}[t]
  \centering
  \footnotesize
  \caption{Retention of the T3.2 validation reward, think and no-think pairs, one training seed per run (exploratory; validation set, 80 evaluations per run, mean over the nine seen and held-out tuples). ``Loss'' is the peak value minus the final value. All runs keep an invalid-output rate of 0.01 or below.}
  \label{tab:t3-retention}
  \begin{tabular}{@{}llcccc@{}}
    \toprule
    & & \multicolumn{2}{c}{Think} & \multicolumn{2}{c}{No-think} \\
    \cmidrule(lr){3-4}\cmidrule(lr){5-6}
    Backbone & Optimizer & peak step & loss & peak step & loss \\
    \midrule
    Qwen3-4B-Instruct   & GRPO   & 800 & 0.00 & 200 & 1.32 \\
    Qwen3-4B-Instruct   & DrGRPO & 800 & 0.00 & 410 & 0.33 \\
    Qwen3-4B-Instruct   & PPO    & 500 & 0.02 & 50  & 0.31 \\
    Qwen2.5-3B-Instruct & GRPO   & 740 & 0.02 & 360 & 1.53 \\
    \bottomrule
  \end{tabular}
\end{table}

A Meta-Llama-3.2-3B-Instruct~\citep{grattafiori2024llama3} pair trained under the same protocol (one seed per arm) reverses the ordering. On the test suite its no-think arm leads on four of the five WOFOST tuples (raw think minus no-think $-0.21$ on \texttt{lnpk} profit, $-0.65$ on \texttt{lnpkw} profit, $-0.31$ on \texttt{lnw} profit, $-2.69$ on \texttt{lnpkw} nutrient), think leads only on \texttt{lnw} yield ($+0.37$), and the two arms are level on DSSAT profit ($-0.07$) and Cycles profit (identical). On DSSAT profit both Llama arms score above the Qwen unified think policy: 1.10 (think) and 1.11 (no-think) in normalized return, from their single seed, against $0.81 \pm 0.13$. We therefore state the reasoning result for the Qwen backbones evaluated.

\textbf{Qwen3-8B (exploratory).} In the author-response period we also trained a think and no-think pair of the hybrid-reasoning Qwen3-8B release~\citep{yang2025qwen3} on T3.2, one seed per arm. In training, the no-think arm had persistent format failures: its training invalid-output rate started at 1.00 and averaged 0.83, and the run stopped after 493 of its 800 planned steps. The think arm completed all 800 steps, with its training invalid-output rate falling from 0.30 to 0.09. Only the think arm was scored on the test suite, at its terminal checkpoint: it returns 0.40 on the held-out WOFOST \texttt{lnpkw} nutrient tuple and 0.31 on WOFOST \texttt{lnw} profit in raw objective return (no-action: $-2.87$ and $0.01$), and on both Cycles tuples, scored on the first-build Cycles suites (Appendix~\ref{app:test-suite}), its invalid-output rate is 1.00: it never learned their action format. The pair does not isolate reasoning from format compliance.

\FloatBarrier
\section{Discussion details}
\label{app:discussion-details}

\FloatBarrier
\subsection{Cross-tier summary}
\label{app:cross-tier}

Table~\ref{tab:cross-tier} collects, setting by setting, what the three-seed test-suite results support. It restates Sections~\ref{sec:tier1}--\ref{sec:tier3}, which give the numbers.

\begin{table}[h]
  \centering
  \footnotesize
  \setlength{\tabcolsep}{4pt}
  \renewcommand{\arraystretch}{1.15}
  \caption{When prompt-conditioned LLM-RL generalizes: what the frozen test suite supports at three training seeds per arm, per setting. ``Think'' is the Qwen3-4B-Instruct policy trained with explicit reasoning; ``untrained'' is the same backbone before RL.}
  \label{tab:cross-tier}
  \begin{tabularx}{\linewidth}{@{}l l Y@{}}
    \toprule
    Tier & Setting & Supported finding \\
    \midrule
    1 & weather-regime & Think is above the untrained model in every seed and split and above PPO at matched updates, A2C and DQN (a single-seed PPO with eight gradient passes is higher on nine of ten targets; Appendix~\ref{app:t1-neural}); no-think stays near no-action. \\
    1 & crop-variety & Think is above the untrained model and PPO \texttt{n1} on every coverage and split (PPO with eight gradient passes, three seeds, is ahead at 4ID and 8ID; Appendix~\ref{app:t1-neural}); no-think stays near no-action. \\
    1 & price-regime & Think and no-think have similar means (no observed advantage for reasoning), both above PPO \texttt{n1} and the untrained model. \\
    \addlinespace[0.2em]
    2 & observation-schema & Transfer holds under renamed, anonymized, added, or dropped secondary fields; without the resource state, RL training leaves think below the untrained model on average (one of three seeds above it) in this ablation and training setup. \\
    2 & action-menu & Family~A gains over the untrained model on every test menu in every seed, Family~B only on its own training menu \texttt{lw}; consistent with joint action coverage. \\
    2 & reward-formulation & A think policy trained on yield alone transfers to three unseen objectives, and multi-objective training adds a modest margin; multi-objective no-think training is unstable. \\
    \addlinespace[0.2em]
    3 & cross-simulator & Four of six single-source transfers (both into WOFOST, both into Cycles) score below the untrained model in every seed; the two into DSSAT gain only over an untrained model that fails the output format there; DSSAT is the hardest target. A dynamics limit is one possible explanation, which these runs do not isolate. \\
    3 & unified policy & Trained only on other tuples, the unified think policy learns DSSAT and reaches about 86\% of the gain of the specialist trained on that tuple; the specialists stay ahead on WOFOST \texttt{lnw} profit and Cycles profit, where the unified mean also ends below the untrained model; no-think is unstable across seeds. \\
    \bottomrule
  \end{tabularx}
\end{table}

Results outside this table are exploratory: the training curves (Appendix~\ref{app:full-results-training-curves}) and the backbone comparisons of Appendix~\ref{app:t3-backbones}, which use the validation set or a single training seed, and the crop-variety per-crop breakdown (Appendix~\ref{app:t12-percrop}), which splits the test suite below its prespecified targets. The association of thinking with stable training holds for the Qwen backbones we evaluated; a single-seed Llama-3.2-3B pair reverses it, and in the Qwen3-8B pair the no-think arm has persistent format failures (Appendix~\ref{app:t3-backbones}).

\FloatBarrier
\subsection{Operational costs}
\label{app:operational-costs}

LLM-RL costs more to train than the neural baseline. A single Tier~1 LLM-RL run trains the 4B policy on two H200 GPUs for about eight hours, while PPO on weather-regime with the same interaction budget finishes in about five (chickpea) to nine (potato) hours on a CPU node without a GPU, and the 19-year price-regime rotation takes about 88 hours per seed (W\&B-recorded run time; Appendix~\ref{app:exp-protocol}). Explicit reasoning is the main cost driver on the LLM side, at roughly 215 to 245 generated tokens per decision. Text-to-action generation also adds a failure mode that a numerical policy does not have: an output the environment cannot parse into a legal action (Appendix~\ref{app:t3-details} reports these rates for Tier~3).

\FloatBarrier
\subsection{Limitations in detail}
\label{app:limitations}

\textbf{Simulation only.} DSSAT, WOFOST, and Cycles are mechanistic crop simulators without real sensor noise, actuator latency, or farmer-in-the-loop delays. Our results describe generalization across simulated scenarios and policy-facing schemas, and we make no claim about transfer to real farms; that claim would need field validation.

\textbf{Seeds and exploratory results.} Each arm trains three seeds, and differences between independently trained arms are stated in propagated SD (Appendix~\ref{app:exp-protocol}). Three seeds establish the direction and spread of an effect and are too few for formal significance tests. Confirmatory results are the three-seed test-suite results at the terminal checkpoint; results that rest on the validation set or on one training seed are exploratory (Appendix~\ref{app:cross-tier}). All main results use one backbone, Qwen3-4B-Instruct.

\textbf{Reading the normalized return.} The scripted agronomic reference is a conservative rule set, uncalibrated against experts or optimal control, so $\tilde{R}$ measures distance from codified practice. On 23 of the 24 normalizable WOFOST targets the untrained think model already scores above 1; a score of 5 means five times the gap between no action and that rule set. Weather-regime cannot be normalized and reports the gain over no-action, and scores are compared only within a target (Appendix~\ref{app:exp-normalization}).

\textbf{Cross-simulator confounding.} Each T3.1 transfer changes the transition dynamics together with the observation fields, the action menu and its units, the objective terms, and the site and weather source (Table~\ref{tab:t3-inventory}), so Tier~3 attributes no transfer result to a single factor; the reading in terms of transition dynamics is one plausible explanation. A Cycles scenario is fixed by its weather year at a single site, so each Cycles target rests on 13 physical scenarios (weather years 2004--2016; Appendix~\ref{app:test-suite}). The Cycles yield tuple of T3.2 is saturated, and the untrained think model fails the output format on the DSSAT profit targets and scores as no-action there (Appendix~\ref{app:t3-details}).

\textbf{Evaluation assets.} The test suite was built after submission (Appendix~\ref{app:test-suite}). Its first build of the three Cycles crop-growth suites reused the eight validation years and was rebuilt from unused years. The DSSAT reference values in the submitted version were set by hand and are replaced here by a reconstruction from code (Appendix~\ref{app:bl-dssat}). The Cycles crop-growth reference is the first implementation of our generic scripted policy, since the current implementation has no nitrogen rule on that task and scores exactly as no-action (Appendix~\ref{app:bl-cycles}).

\FloatBarrier
\section{Extended related work}
\label{app:related-work}

This appendix gives the per-thread discussion summarized by the comparison in Table~\ref{tab:related-work-comparison} of Section~\ref{sec:introduction}.

\FloatBarrier
\subsection{Agricultural RL and crop-management simulators}
\label{app:rw-ag-rl}

Agricultural RL has been built on single-simulator gym environments, each binding one crop-growth model to a fixed observation, action, and reward interface. \emph{gym-DSSAT}~\citep{gautron2022gymdssat} runs on DSSAT, \emph{WOFOSTGym}~\citep{solow2025wofostgym} on WOFOST for annual and perennial crop management, \emph{CyclesGym}~\citep{turchetta2022cyclesgym} on Cycles for long-horizon multi-crop nitrogen scheduling, and \emph{CropGym}~\citep{overweg2021cropgym} on PCSE for nitrogen management. Each binds its own interface to its platform.

Recent work brings language models into agricultural decision-making. \citet{wu2024newagronomists} convert state variables into natural-language text and let a language model encode them for a downstream DQN, and CROPS~\citep{wu2024crops} trains a language model as a DQN agent on partially masked states, both on DSSAT alone; \citet{kaushik2026selfadaptive} compare zero-shot LLM agents, without RL training, against RL agents on agricultural tasks under shifted weather; \citet{chen2024llmrlcropsurvey} survey integrations of RL and LLMs in crop management. None of these trains an LLM policy with RL across multiple simulators, and none evaluates structural variation. The AgriManager framework places a prompt-based schema layer over three otherwise incompatible simulators (WOFOST, DSSAT, Cycles), and the benchmark treats structural variation, formalized as schema shift, as an evaluation axis (Sections~\ref{sec:framework} and~\ref{sec:benchmark}).

\FloatBarrier
\subsection{RL generalization and cross-task transfer}
\label{app:rw-rl-gen}

Generalization in reinforcement learning has been studied along several lines, most under the assumption that the policy interface remains fixed between training and deployment. \emph{Contextual MDPs} and procedural-content benchmarks~\citep{hallak2015contextual,cobbe2020procgen} hold the interface fixed and vary a latent context such as a random seed, task identifier, or parameter vector. The survey of \citet{kirk2023survey} catalogs this zero-shot literature, in which the observation, action and reward schema remain invariant. Sim-to-real transfer and \emph{domain randomization}~\citep{tobin2017dr,zhao2020sim2real} vary the dynamics $P$ and observation-rendering parameters to close the reality gap between simulated and physical platforms; the interface shape stays fixed.

Transfer RL has also studied changes in parts of the interface. \citet{zhu2023transfersurvey} survey transfer across differences in MDP components, including the state, action and reward spaces. \citet{sun2022obsfeature} transfer across observation feature spaces with the action and reward schema fixed, and \citet{sivakumar2024heterogeneous} transfer across heterogeneous action spaces through subgoal mapping. For LLM agents, \citet{xi2026rlgeneralization} study whether RL improves generalization to environments with different observation spaces and action interfaces, OpenAgent~\citep{wu2026openagent} varies the query, action, observation and domain of tool-use agents, and \citet{liu2026actionspace} show that tool-use performance depends on how a tool's schema is written. These works vary one or two interface components, or study tool-calling agents. \emph{Schema shift} (Section~\ref{sec:framework-pomdp}) varies the observation, action, and reward schema of sequential control tasks, separately and jointly, and the AgriManager benchmark evaluates each axis and their composition in one ladder.

\FloatBarrier
\subsection{Prompt-conditioned LLM policies and agent benchmarks}
\label{app:rw-llm-agents}

Prompt-conditioned LLM policies and LLM-agent benchmarks make interface variation expressible at deployment, yet their evaluations do not target multi-turn RL across heterogeneous decision schemas. \emph{LLaRP}~\citep{szot2024llarp} adapts a frozen LLM into an embodied policy by training its input and action modules with RL; \emph{SayCan}~\citep{ahn2022saycan} couples LLM reasoning with affordance value functions for grounded robot control; \emph{ReAct}~\citep{yao2023react} interleaves reasoning traces and actions in the model's generated output. All three demonstrate that the policy can read its task specification from a natural-language prompt, but evaluation stays within a single decision schema.

Two LLM-agent benchmarks span heterogeneous tasks: \emph{AgentBoard}~\citep{ma2024agentboard} covers nine task domains, and \emph{WebArena}~\citep{zhou2024webarena} provides 812 long-horizon web tasks; both evaluate prompted agents without RL training. Their variation is in tool use and web navigation; the decision schema within each task remains fixed. Among RL-trained LLM agents, \citet{xi2026rlgeneralization} vary observation spaces and action interfaces across environments; we know of no benchmark that also varies the objective and evaluates these axes separately and jointly in agricultural decision-making, the regime the AgriManager benchmark targets.

\end{document}